\documentclass{article} 
\usepackage[preprint]{colm2025_conference}

\usepackage{microtype}
\usepackage{hyperref}
\usepackage{url}
\usepackage{booktabs}
\usepackage{lineno}
\usepackage{inconsolata}
\usepackage{paralist}
\usepackage{soul}
\usepackage{graphicx}
\usepackage{amsmath}
\usepackage{amsthm}
\usepackage{algorithm}
\usepackage{algpseudocode} 
\usepackage{nicefrac}
\usepackage{xcolor}
\usepackage{cleveref}
\usepackage{array}
\usepackage{longtable}
\usepackage{enumitem}
\usepackage{amssymb}
\usepackage{xspace}
\usepackage{makecell}
\usepackage{bbding}
\usepackage{boldline}
\usepackage{pifont}
\usepackage{tabularx}
\usepackage{wrapfig}
\usepackage[export]{adjustbox}
\usepackage{fvextra}
\usepackage[most,breakable]{tcolorbox}
\usepackage{caption}
\usepackage{pgfplots}
\usepackage{multirow}
\usepackage{subcaption}

\usepackage{booktabs}
\usepackage{multirow}
\usepackage{makecell}
\usepackage{arydshln}
\usepackage{graphicx}

\pgfplotsset{compat=1.18}

\usepackage[commandnameprefix=always]{changes}
\usepackage{lipsum}

\definecolor{darkblue}{rgb}{0, 0, 0.5}
\hypersetup{colorlinks=true, citecolor=darkblue, linkcolor=darkblue, urlcolor=darkblue}

\newcommand{\ours}[0]{Reasoning-Table\xspace}
\newcommand{\think}[1]{\textcolor{blue}{\texttt{<think>}} #1 \textcolor{blue}{\texttt{</think>}}}

\newcommand{\answer}[1]{\textcolor{purple}{\texttt{<answer>}} #1 \textcolor{purple}{\texttt{</answer>}}}
\newcommand{\position}[2]{\textcolor{orange}{\texttt{<\textbar{}#1\textbar{}><\textbar{}#2\textbar{}>}}}
\newcommand{\oneposition}[2]{\textcolor{orange}{\texttt{<\textbar{}#1\textbar{}>}}}

\newcommand{\hf}{\raisebox{-1.5pt}{\includegraphics[height=1.05em]{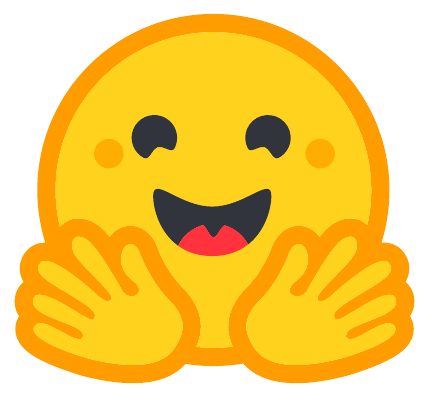}}\xspace}
\newcommand{\git}{\raisebox{-1.5pt}{\includegraphics[height=1.05em]{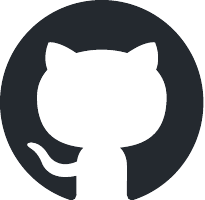}}\xspace}

\title{Reasoning-Table: Exploring Reinforcement Learning for Table Reasoning 
}

\author{Fangyu Lei$^{1,2}$\thanks{Equal contribution. Authors in alphabetical order.}~,
Jinxiang Meng$^{1,2*}$, 
Yiming Huang, \\
\textbf{Tinghong Chen}$^2$,
\textbf{Yun Zhang}$^1$, 
\textbf{Shizhu He}$^{1,2}$, 
\textbf{Jun Zhao}$^{1,2}$, 
\textbf{Kang Liu}$^{1,2}$ \\
$^1$ Institute of Automation, CAS ~~
$^2$ University of Chinese Academy of Sciences \\ \\
\hf Datasets: \url{https://huggingface.co/datasets/TableQAKit/Reasoning-Table} \\
\git Code: \url{https://github.com/MJinXiang/Reasoning-Table}
}

\begin{document}

\maketitle

\begin{abstract}

Table reasoning, encompassing tasks such as table question answering, fact verification, and text-to-SQL, requires precise understanding of structured tabular data, coupled with numerical computation and code manipulation for effective inference. 
Supervised fine-tuning (SFT) approaches have achieved notable success but often struggle with generalization and robustness due to biases inherent in imitative learning. 
We introduce Reasoning-Table, the first application of reinforcement learning (RL) to table reasoning, achieving state-of-the-art performance. 
Through rigorous data preprocessing, reward design, and tailored training strategies, our method leverages simple rule-based outcome rewards to outperform SFT across multiple benchmarks. 
Unified training across diverse tasks enables Reasoning-Table to emerge as a robust table reasoning large language model, surpassing larger proprietary models like Claude-3.7-Sonnet by 4.0\% on table reasoning benchmarks. 
The approach also achieves excellent performance on text-to-SQL tasks, reaching 68.3\% performance on the BIRD dev dataset with a 7B model.
Further experiments demonstrate that Reasoning-Table enhances the model's generalization capabilities and robustness.

\end{abstract}

\section{Introduction}

Tabular data provide a foundational structure for storing, organizing, and presenting information, underpinning data analysis and data engineering, which are essential in fields like business intelligence and scientific research \citep{borisov2022tabulardata}.

Table reasoning, recognized as a critical task, involves extracting insights, performing logical inferences, and generating precise responses from structured tabular data~\citep{wikitablequestion,zhu2021tat}. 
This task involves various task types, such as table question answering \citep{wikisql,wikitablequestion}, fact verification \citep{tabfact}, and text-to-SQL \citep{li2024bird,spider2}, and handles various table formats. 
The advent of Large Language Models (LLMs) \citep{gpt4,anthropic2024claude} has markedly enhanced table reasoning capabilities, with cutting-edge frameworks exhibiting an increasing reliance on these models.

Prompt engineering crafts task-specific pipelines for table reasoning \citep{cheng2022binder, chain-of-table, ye2023table, api-table, deng2025reforce}, achieving strong performance on targeted tasks. However, it struggles with scalability and generalizability, requiring significant manual effort.
Alternatively, supervised fine-tuning (SFT) enhances LLMs' understanding of tables \cite{zhang-etal-2024-tablellama, zha2023tablegpt, zhang2024tablellm}, yet struggles with generalization and adaptability due to imitation learning biases~\cite{chu2025sft, guo2025deepseek}, often underperforming in tasks requiring complex reasoning, novel table structures, or unseen task types, leading to poor performance.

\begin{figure*}[t]
    \centering
    \includegraphics[width=\textwidth]{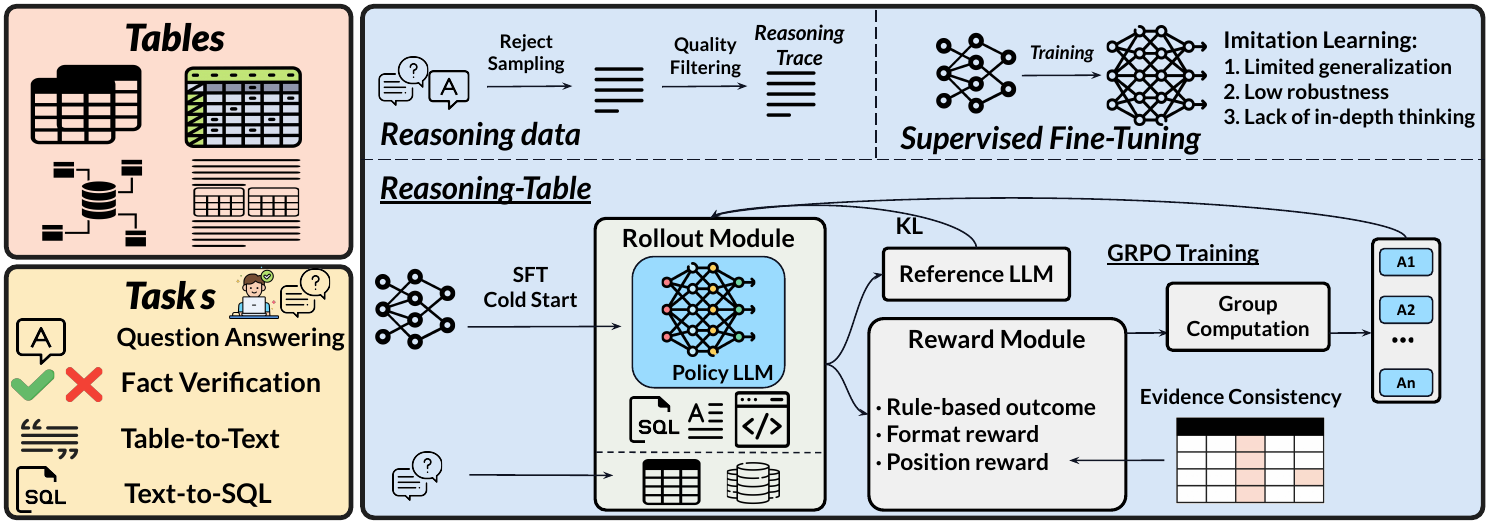}
    \caption{\ours encompasses all table-related tasks and diverse table types. Compared to SFT, using the GPRO RL method, the reasoning model achieves stronger reasoning capabilities, generalization, and robustness.}
    \label{fig:main}
\end{figure*}

Recent advanced LLMs \citep{jaech2024openaio1,anthropic2024claude,guo2025deepseekr1} demonstrate that deeper reasoning via increased computation improves response quality. This has driven significant RL-based progress in domains like mathematical reasoning \citep{yu2025dapo} and code generation \citep{deepcoder2025}.
RL offers the potential to acquire high-quality reasoning paths through independent exploration of the solution space, instead of relying solely on teacher model distillation. This self-directed learning can help mitigate bias and allows for a deeper discovery of effective solution patterns. We explore applying RL to table reasoning tasks, aiming to enhance LLMs' capability in handling complex tables and questions by training deep thinking models without needing to distill from proprietary sources.

In this work, we introduce \ours, the first LLM reinforcement learning method specifically designed for almost all table reasoning tasks. 
We develop a comprehensive pipeline encompassing dataset generation, data filtering strategies, and reinforcement training methods. 
\ours leverages RL to significantly boost performance on diverse table reasoning tasks, including table question answering (tableQA), table fact verification (tableFV), table-to-text generation, and text-to-SQL tasks.
\ours, which utilizes simple rule-based output reward and requires neither costly reasoning trace distillation nor as much data as SFT, substantially surpasses the pure SFT baseline by an average of 17.36\%.
We also propose table-specific reward mechanisms to better leverage the table structure, particularly a position evidence consistency reward that addresses the reward sparsity inherent in RL paradigms applied to table and database interactions.
Notably, unified reinforcement learning training across $7$ tableQA datasets enables \ours to surpass larger proprietary models, such as Claude-3.7-Sonnet, showcasing outstanding effectiveness and robustness.
Training on text-to-SQL datasets also demonstrates excellent performance, achieving a strong result of 68.3\% on the BIRD SQL leaderboard among models of the same size using a 7B model.

Further experiments demonstrate the superior potential of RL training over SFT for table reasoning tasks, as shown by more convenient training data acquisition, enhanced performance, stronger generalization capabilities, and improved robustness under perturbations. 
This superiority stems from RL's ability to improve reasoning processes through autonomous exploration and optimization based on reward signals, rather than being limited by biases inherited from teacher models.
Extensive empirical analyses reveal valuable insights into table-specific reward design, unified task training strategies, and data selection for table reasoning.
All our curated high-quality datasets, training configurations, and implementations will be publicly released, laying a solid foundation for future research and scaling in table reasoning tasks.

\newpage

\section{Table Reasoning}
\label{sec:table_reasoning}

\textbf{Task types.}
Table \ref{tab:datasets} summarizes the $15$ datasets we used for table reasoning. The tasks are primarily divided into $4$ major categories: tableQA, tableFV, table-to-text and text-to-SQL. Text-to-SQL is a significant branch of tableQA, classified separately due to its distinct characteristics.

\textbf{Task formulation.}
Given a natural language query $Q$ and table(s) $\mathcal{T} = \{T_1, T_2, \dots, T_n\}$ ($n \ge 1$), the goal is to derive an answer $A$. In tableQA, $A$ is a string or number extracted from $\mathcal{T}$. In tableFV, $A$ is a boolean ($1$ or $0$) that verifies $Q$ against $\mathcal{T}$. In table-to-text, $A$ is a text sequence generated from $\mathcal{T}$. In text-to-SQL, $A$ is a SQL query derived from $Q$ to retrieve or manipulate data from database $\mathcal{T}$.

\textbf{Table types.}
Tables are diverse and include:
(1) \textit{Simple Column-Header Tables}: Basic CSV- or HTML-like tables with only column headers.
(2) \textit{Spreadsheet Tables}: Feature hierarchical or multi-level headers, row and column headers, and mixed data types (text, numbers, formulas).
(3) \textit{Database Tables}: Stored in relational databases, often with multiple interrelated tables, foreign keys, and indexes.
(4) \textit{Embedded/Semi-Structured Tables}: Found in documents (e.g., PDFs, Word, webpages), with text interspersed around or within the table.

\section{Methodology}
\label{sec:data_process}

\subsection{SFT Dataset Preprocessing}
\label{sec:sft_preprocess}
Most table reasoning datasets lack annotations for rationales. We generate two types of SFT data: \textit{No-Reason SFT} and \textit{Reason-SFT}.

\textbf{No-Reason SFT data.}
We use the full original datasets to construct No-Reason SFT, with the input data format as shown in Table \ref{tab:reason_sft_generation}. The total data volume is provided in Table \ref{tab:filtered_datasets}.

\textbf{Reason-SFT data sampling.}
To create high-quality long chain-of-thought data, we use proprietary LLMs \citep{guo2025deepseekr1, anthropic2024claude, openai2023gpt4} to generate reasoning traces from existing table reasoning datasets.
Each example is processed $5$ times to produce multiple outputs. For tableQA and tableFV, we retain instances with correct responses based on the EM metric. For table-to-text generation tasks, we use the BLEU score as the evaluation criterion, keeping samples with scores above 0.3. For text-to-SQL tasks, we select trajectories from samples with an execution accuracy of 1. The specific sample counts can be found in Table~\ref{tab:filtered_datasets}.
We found this approach to be costly and dependent on LLM performance, and we cannot generate reasoning traces for the entire dataset, as model performance cannot reach 100\%.

\textbf{Reason-SFT data quality filtering.}  
To ensure high-quality SFT data, a redundancy detection algorithm refines reasoning processes within \texttt{<think>} tags from Deepseek-R1. The process: (1) segments text into sentences, discarding those with fewer than five words; (2) computes TF-IDF-based cosine similarity; (3) adjusts similarity scores with penalties for question mismatch and modal verb disparity; and (4) removes examples with more than two sentence pairs exceeding a similarity threshold ($\tau = 0.7$). This filters out approximately 90.3\% of trajectories, producing refined datasets detailed in Table~\ref{tab:filtered_datasets}, with further details in App.\ref{app:reason_sft_details}.

\subsection{RL Dataset Preprocessing}
\label{sec:rl_preprocess}
\textbf{Difficulty control.}
For table reasoning tasks, RL excels because dataset answers are carefully human-annotated, unlike LLM-generated rationales. Notably, controlling QA pair alignment and data difficulty is also essential. We used Qwen2.5-32B to generate $8$ answers per example, with results reported in Table~\ref{tab:rollout_datasets}. The results are used for subsequent data ablation study (Sec.\ref{sec:data_ablation}). Initially, we used the full dataset.

\textbf{Position evidence.}
The column-row structure is a distinguishing feature of tabular data compared to other data types. To leverage this, we introduce a table-specific reasoning content called \emph{position trace}. 
We employ a prompting strategy (see Table~\ref{tab:position_instruction}) with two modes: \textit{(1) Column localization}: Identifying relevant columns. \textit{(2) Column and Cell localization}: Identifying both columns and specific cells.
For position annotation, we use LLMs to sample each example 5 times. We filter for samples with correct final answers and compute the union of their position annotations, i.e.,
$\bigcap_{i=1}^5 P_i$,
where $ P_i $ denotes the position set from the $ i $-th correct sample. This ensures robust and comprehensive position evidence.
We extract the position evidence for three datasets, 
with 71.17\% yielding valid evidence,
statistic details are shown in App.\ref{sec:sft_data_quailty}. 
For some datasets where the rows and columns containing the answers have already been annotated, we directly utilize this information \citep{zhu2021tat,cheng2022hitab}.
For text-to-SQL tasks, we use the column names that appear in the gold SQL as the position reward, which is also commonly referred to as schema linking \citep{yu2018spider,li2024bird}.




\subsection{Training}

\textbf{Supervised fine-tuning.}
Let \(\{(x_i, y_i)\}_{i=1}^N\) be a labeled dataset of input-target pairs. We seek to optimize the model parameters \(\theta\) by minimizing the standard cross-entropy loss:
\[
\mathcal{L}_{\text{SFT}}(\theta) 
= -\sum_{i=1}^{N} \log \pi_\theta\bigl(y_i \mid x_i\bigr),
\]
where \(\pi_\theta(\cdot \mid x)\) denotes the model’s predicted probability distribution for the target given the input \(x\).

\textbf{Reinforcement training.}
Inspired by recent advanced reasoning models, We use GRPO to train the reasoning model. GRPO uses relative performance within a group of candidates to compute an advantage $A_i$ for policy updates based on the provided objective function.

\begin{small}
\begin{equation*}
\begin{split}
\mathcal{J}_{\text{GRPO}}(\theta) = \mathbb{E}_{\mathbf{x} \sim P(\mathbf{X})} \mathbb{E}_{o_i \sim \pi_{\text{old}}} \Big[ 
\frac{1}{G} \sum_{i=1}^{G} \Big( 
\min\Big( \frac{\pi_\theta(o_i|\mathbf{x})}{\pi_{\text{old}}(o_i|\mathbf{x})} A_i, \\
\text{clip} \big( \frac{\pi_\theta(o_i|\mathbf{x})}{\pi_{\text{old}}(o_i|\mathbf{x})}, 1-\epsilon, 1+\epsilon \big) A_i 
\Big) - \beta D_{\text{KL}}(\pi_\theta \| \pi_{\text{ref}})
\Big) \Big]
\end{split}
\end{equation*}
\end{small}
where \( r_i^{\text{ratio}} = \frac{\pi_\theta(o_i|\mathbf{x})}{\pi_{\text{old}}(o_i|\mathbf{x})} \) is the importance sampling ratio, and \( A_i \) denotes the group-relative advantage. The hyperparameters \( \epsilon \) and \( \beta \) control the update range and divergence from the reference policy.

\subsection{Reward Design}

\textbf{Rule-based outcome reward.}~($R_{\text{ans}}$).
The model's final answer must be enclosed within an \texttt{<answer>} tag.
The reward function evaluates the predicted answer \( A \) against the reference answer \( A_{\text{ref}} \), producing a binary reward \( R_{\text{ans}} \in \{0, 1\} \), where 1 indicates a correct or equivalent answer, and 0 indicates an incorrect answer.
The rule-based reward uses $4$ evaluation metrics:
\textit{(1) Exact match}: Checks if \( A = A_{\text{ref}} \).
\textit{(2) Execution result match}: For tasks like SQL, checks if \( \text{Exec}(A) = \text{Exec}(A_{\text{ref}}) \).
\textit{(3) F1 thresholding}: Assigns 1 if the token-level F1 score \( \text{F1}(A, A_{\text{ref}}) \geq \phi \), where \( \phi \) is a threshold, and 0 otherwise.
\textit{(4) BLEU thresholding}: Assigns 1 if \( \text{BLEU}(A, A_{\text{ref}}) \geq \tau \), where \( \tau \) is a threshold, and 0 otherwise.
Calculation details are provided in App.\ref{app:metrics}.
The rule-based reward is defined as:
\[
R_{\text{ans}} =
\begin{cases}
1 & \text{if } A = A_{\text{ref}} \\
  & \textit{ (Short string or number answer)} \\
1 & \text{if } \text{F1}(A, A_{\text{ref}}) \geq \phi \\ 
  &  \textit{ (Long string answer)} \\
1 & \text{if } \text{Exec}(A) = \text{Exec}(A_{\text{ref}}) \\
  & \textit{(Text-to-SQL)} \\
1 & \text{if } \text{BLEU}(A, A_{\text{ref}}) \geq \tau \\ 
  &  \textit{ (Long-Form QA, Table-to-text)} \\
0 & \text{otherwise}
\end{cases}
\]

\textbf{Format reward}~($R_{\text{fmt}}$).
To ensure structural consistency and output readability, we enforce the presence of both \texttt{<think>} and \texttt{<answer>} tags in the model's output.

\textbf{Position reward}~($R_{\text{pos}}$).
To encourage faithful reasoning grounded in the table, the model is required to annotate table cells with \texttt{<|cell content|><|column name|>} tags when directly referencing them in reasoning. Each annotated cell provides evidence for the reasoning process. The position reward evaluates how well the model identifies and annotates the relevant cells during reasoning:$R_{\text{pos}} = \frac{|P \cap G|}{|P\cup G|}$
where $P$ denotes the set of cells annotated by the model and $G$ represents the ground-truth set of cells that should be referenced to answer the question. This encourages models to precisely identify the relevant information in the table rather than over-annotating or missing critical values.
The model's final result must be consistent with its predicted position. This reward function ensures that the final answer is directly derived from the model's attention positions, rather than being generated in isolation from the reasoning steps.

\textbf{Final reward}~($R$).
Simply summing these rewards could allow the model to achieve a high score by prioritizing consistency (e.g., correct cell annotations or format) even when the answer is incorrect. To address this, we design the final reward to prioritize answer correctness while rewarding proper reasoning and formatting:
\[
R(o_i) = R_{\text{ans}}(o_i) \times (1 + \lambda_1 R_{\text{pos}}(o_i)) + \lambda_2 R_{\text{fmt}}(o_i)
\]
where $\lambda_1$ and $\lambda_2$ are tunable hyperparameters. The multiplication of $R_{\text{ans}}$ with the position reward term ensures that $R_{\text{pos}}$ contributes only when the answer is correct, preventing the model from being overly rewarded for consistent but incorrect reasoning. The format reward, weighted by $\lambda_2$, encourages structural compliance independently.

\vspace{-5pt}
\section{Experiments}
\label{sec:experiments}

\subsection{Setup}
\textbf{Benchmarks.}
We use several table reasoning datasets, covering tableQA, tableFV, table-to-text, and text-to-SQL, with format prompt for inference (See Tab.\ref{tab:train_prompt}). 
We compare our method against large open-source LLMs, proprietary LLMs (Claude-3.7-Sonnet, GPT-4o), and the table-specific domain model TableGPT2 \citep{su2024tablegpt2}.

\textbf{Single dataset training settings.}
We use Qwen2.5-7B models as the base model.
We employ four training settings for each dataset:
\textit{(1) No-Reason SFT}: 
We use the input and output data from the dataset to train the model, but not all data includes high-quality reasoning traces.
\textit{(2) Reason-SFT Start}: 
We use the chain-of-thought reasoning traces generated through rejection sampling (Sec.\ref{sec:sft_preprocess}) to train the model.
\textit{(3) RL-zero}: We use the GRPO algorithm \cite{shao2024deepseekmath} for RL on the data. The learning rate is $1e-6$, rollout number is 16, batch size is 32, and epochs are 3. We set $\beta = 1e-3$ for KL regularization and apply a generation temperature of 1.0. We utilize the rule-based outcome reward, set $\lambda_1=0, \lambda_2=0.2$.
\textit{(4) Reason SFT+RL}: We use Reason-SFT data for a cold start to initialize the model to learn the reasoning trace format, followed by GRPO training on the fine-tuned checkpoints.

\textbf{Unified training settings.}
Validating performance on individual dataset tasks is insufficient for assessing the method’s comprehensive capabilities, making unified training critical. 
We conduct unified fine-tuning and reinforcement learning across all tableQA datasets. To address excessively large training sets in some datasets, we adjust data quantities to balance the model’s reward, with details provided in Table~\ref{tab:filtered_datasets}.
For SFT, we use a batch size of 64, a learning rate of 5e-6, and train for 5 epochs. For RL, we employ a learning rate of 1e-6, utilize rule-based rewards, and set $\lambda_1 = 0$ and $\lambda_2 = 0.2$, indicating that both answer reward and format reward are used simultaneously.

\textbf{Metrics.}
We use Exact Match(EM) for tableQA and tableFV, Execution Accuracy (EX) for text-to-SQL, BLEU for table-to-text task (Sec.\ref{app:metrics}).
For the base model and the unified training model, we also use the arithmetic mean score to evaluate the overall performance on the tableQA task.
Details are provided in App.\ref{sec:exp_details}.

\begin{table*}[htbp]
\centering
\caption{Performance on tableQA tasks. The metric is EM or Accuracy. AVG is the arithmetic mean of seven tasks. We only use the train set for training and the test set for evaluation, without using the validation set.}
\resizebox{0.99\textwidth}{!}{
\begin{tabular}{lcccccccc}
\toprule
\textbf{Method} & \textbf{WikiTQ} & \textbf{HybridQA} & \textbf{MultiHiertt} & \textbf{OTT-QA} & \textbf{FinQA}  & \textbf{TAT-QA} & \textbf{HiTab} & \textbf{AVG}  \\
\midrule
Claude & 82.02 & 39.36 & 40.98 & 62.69 & 57.45 & 53.09 & 75.96 & 58.79 \\
GPT-4o & 81.19 & 39.30 & 40.86 & 66.35 & 57.63 & 53.45 & 73.92 & 58.96 \\
TableGPT2-7B & 63.70 & 30.03 & 25.12 & 48.87 & 38.36 & 55.12 & 63.89 & 46.44 \\
Qwen2.5-32B-Inst & 79.65 & 38.2 & 37.74 & 56.5 & 59.2 & 67.29 & 73.29 & 58.84 \\ 
Qwen2.5-7B-Inst & 57.27 & 31.84 & 27.54 & 50.50 & 52.40 & 49.79 & 57.19 & 46.65 \\
\midrule
\multicolumn{9}{c}{\textbf{Single Dataset Training}} \\ \midrule
No-Reason SFT & 58.19 & 36.35 & 15.38 & 42.52 & 57.36 & 54.46 & 78.28 & / \\
Reason-SFT & 74.88 & 37.18 & 36.61 & 55.94 & 63.38 & 71.59 & 74.43 & / \\
RL-zero & 76.35 & 43.68 & 37.80 & 65.13 & 66.87 & 74.38 & 80.56 & / \\
Reason-SFT+RL & \textbf{80.07} & \textbf{44.64} & \textbf{40.97} & \textbf{68.27} & \textbf{68.32} & \textbf{79.79} & \textbf{81.94} & / \\ 
\midrule
\multicolumn{9}{c}{\textbf{Unified Training}} \\ \midrule
No-Reason SFT & 57.21 & 42.36 & 15.92 & 42.44 & 58.41 & 53.57 & 78.72 & 49.80 \\
Reason-SFT & 72.35 & 35.17 & 38.50 & 54.40 & 60.42 & 63.45 & 72.72 & 56.72 \\
RL-zero & 72.75 & 41.37 & 37.23 & 67.67 & 60.41 & 69.87 & 71.53 & 60.12 \\
Reason SFT+RL & 75.46 & 42.83 & 39.56 & 68.68 & 64.46 & 73.75 & 73.61 & 62.62 \\
\bottomrule
\end{tabular}
}
\label{tab:tqa_result}
\end{table*}

\begin{table*}[h]
\centering
\caption{Performance on text-to-SQL (using Qwen2.5-Coder), long-form QA, tableFV and table-to-text tasks.}
\resizebox{0.99\textwidth}{!}{
\begin{tabular}{l|c|cc|c|cc}
\toprule
\multirow{2}{*}{\textbf{Method}} & \textbf{Long-form QA} & \multicolumn{2}{c|}{\textbf{Fact Verification}} & \textbf{Table-to-Text} & \multicolumn{2}{c}{\textbf{Text-to-SQL}} \\
\cmidrule(lr){2-2} \cmidrule(lr){3-4} \cmidrule(lr){5-5} \cmidrule(lr){6-7}
 & \textbf{FetaQA (BLEU)} & \textbf{TabFact (Acc)} & \textbf{FEVEROUS (Acc)} & \textbf{ToTTo (BLEU)} & \textbf{Spider (EX)} & \textbf{BIRD (EX)} \\
\midrule
Claude-3.7-Sonnet & 25.57 & 91.12 & 71.84 & 15.90 & 80.66 & 59.13 \\
GPT-4o & 28.47 & 92.09 & 78.31 & 17.19 & 77.95 & 63.56 \\
TableGPT2-7B & 13.80 & 83.92 & 68.57 & 13.20 & 75.53 & 47.39 \\
Qwen2.5-7B-Inst & 24.96 & 82.82 & 63.62 & 13.92 & 77.75 & 50.72 \\
Qwen2.5-32B-Inst & 27.75 & 89.48 & 74.26 & 14.46 & 84.91 & 64.41  \\ 
\midrule
\multicolumn{7}{c}{\textbf{Single Dataset Training}} \\ \midrule
No-Reason SFT  & 36.08 & 88.15 & 86.84 & 16.30 & 79.13 & 55.38 \\ 
Reason-SFT  & 37.81 & 90.18 & 79.48 & 19.55 & 82.92 & 59.58 \\ 
RL-zero  & 42.96 & 91.46 & \textbf{86.53} & 23.71 & 83.18 & 62.41 \\ 
Reason SFT+RL  & \textbf{43.18} & \textbf{92.77} & 85.67 & \textbf{28.39} & \textbf{84.20} & \textbf{63.16} \\ 
\bottomrule
\end{tabular}
}
\label{tab:other_tasks}
\end{table*}

\subsection{Main Results}
\textbf{RL training exhibiting superior results compared to SFT across all datasets.}
As shown in Table~\ref{tab:tqa_result} and Table~\ref{tab:other_tasks}, we conduct experiments with both SFT and RL across all datasets. 
In single-dataset settings, \textit{RL-zero} consistently outperforms \textit{Reason-SFT} and \textit{No-Reason SFT}. 
In the unified training setting, \textit{RL-zero} surpasses \textit{Reason-SFT} by 10.0\% and \textit{No-Reason SFT} by 3.1\%, while using less data and requiring no reasoning traces from a teacher model. 
This demonstrates RL’s strong exploration capability, enabling the model to learn effectively from its own rollouts for enhanced reasoning performance.

\textbf{SFT cold start enhances RL performance across all tasks.} 
Our experiments demonstrate that initializing with SFT before RL yields significant performance improvements across nearly all tasks. In the unified training setting, \textit{Reason-SFT+RL} outperforms \textit{RL-zero} by an average of 2.79\%, underscoring the critical role of SFT initialization in model training.
Furthermore, on the TableQA task, \textit{Reason-SFT+RL} surpasses Claude-3.5-Sonnet and Qwen2.5-32B by approximately 2.5\%, highlighting the potential of Reasoning-Table’s data and methods for scaling training.

\textbf{Reasoning-intensive tasks benefit more from Reasoning-Table.}
More challenging tasks, such as BIRD text-to-SQL, exhibit greater performance improvements from our methods, whereas tasks like table-to-text and tableFV show smaller gains. On the TabFact dataset, \textit{RL-zero} outperforms \textit{No-Reason SFT} by only 3\%, reflecting the influence of task difficulty on performance gains.

\begin{figure}[t]
    \centering
    \begin{subfigure}[t]{0.32\textwidth}
        \centering
        \includegraphics[width=\textwidth]{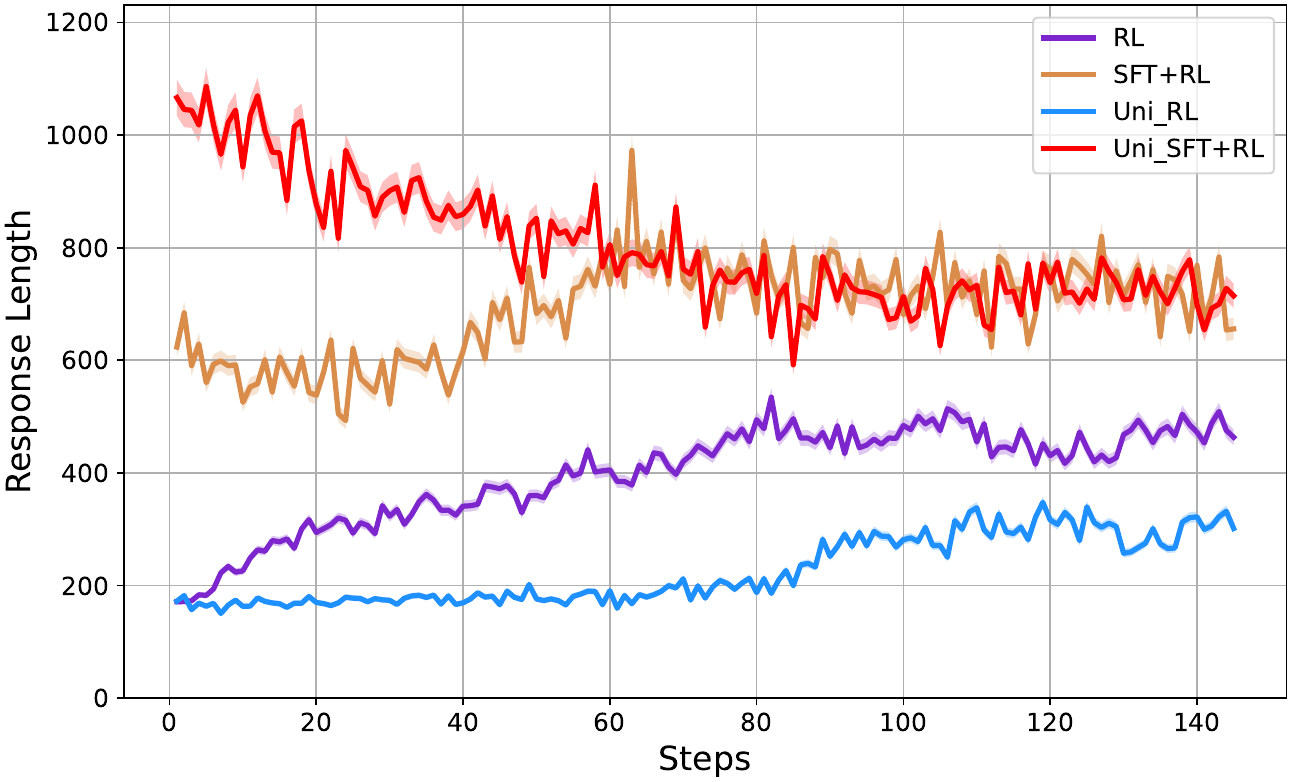}
        \caption{Response length with training steps.}
        \label{fig:wikitq_length_training}
    \end{subfigure}
    \hfill
    \begin{subfigure}[t]{0.32\textwidth}
        \centering
        \includegraphics[width=\textwidth]{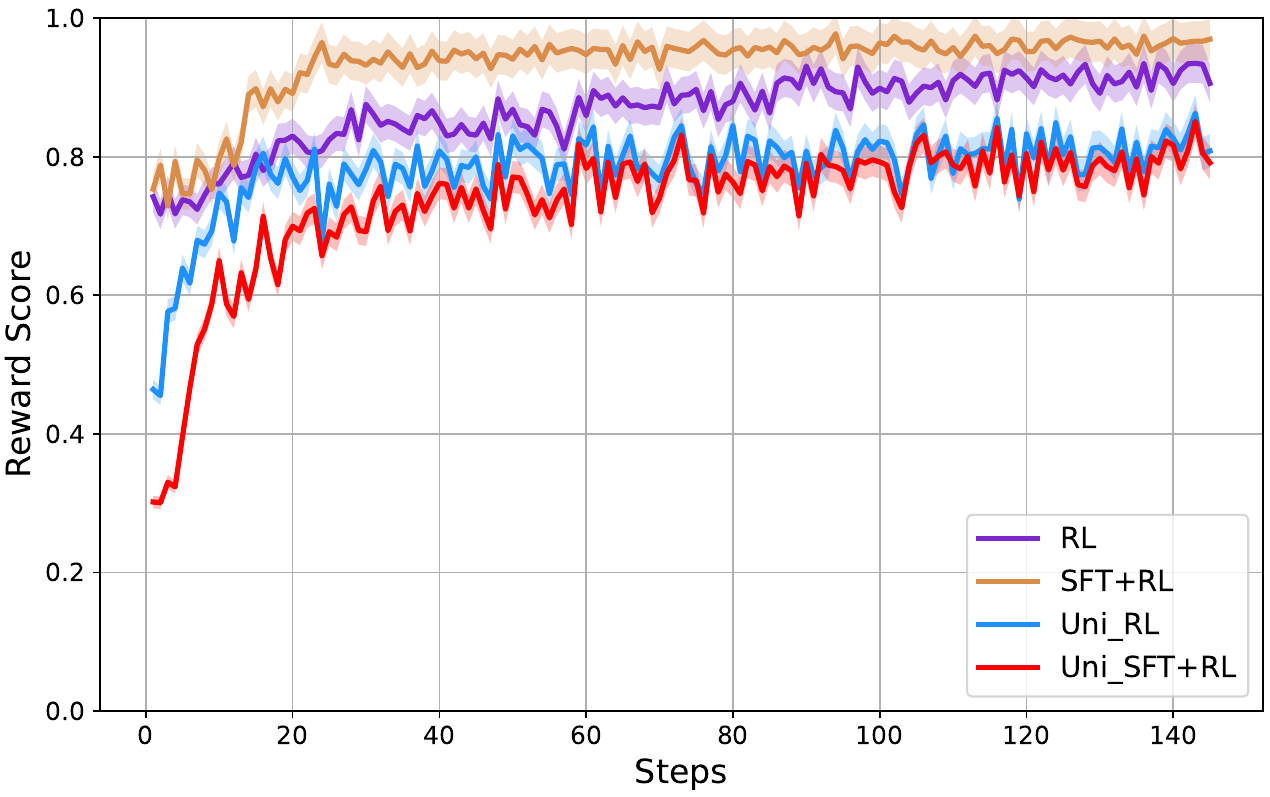}
        \caption{Reward with training steps.}
        \label{fig:wikitq_reward_training}
    \end{subfigure}
    \hfill
    \begin{subfigure}[t]{0.32\textwidth}
        \centering
        \includegraphics[width=\textwidth]{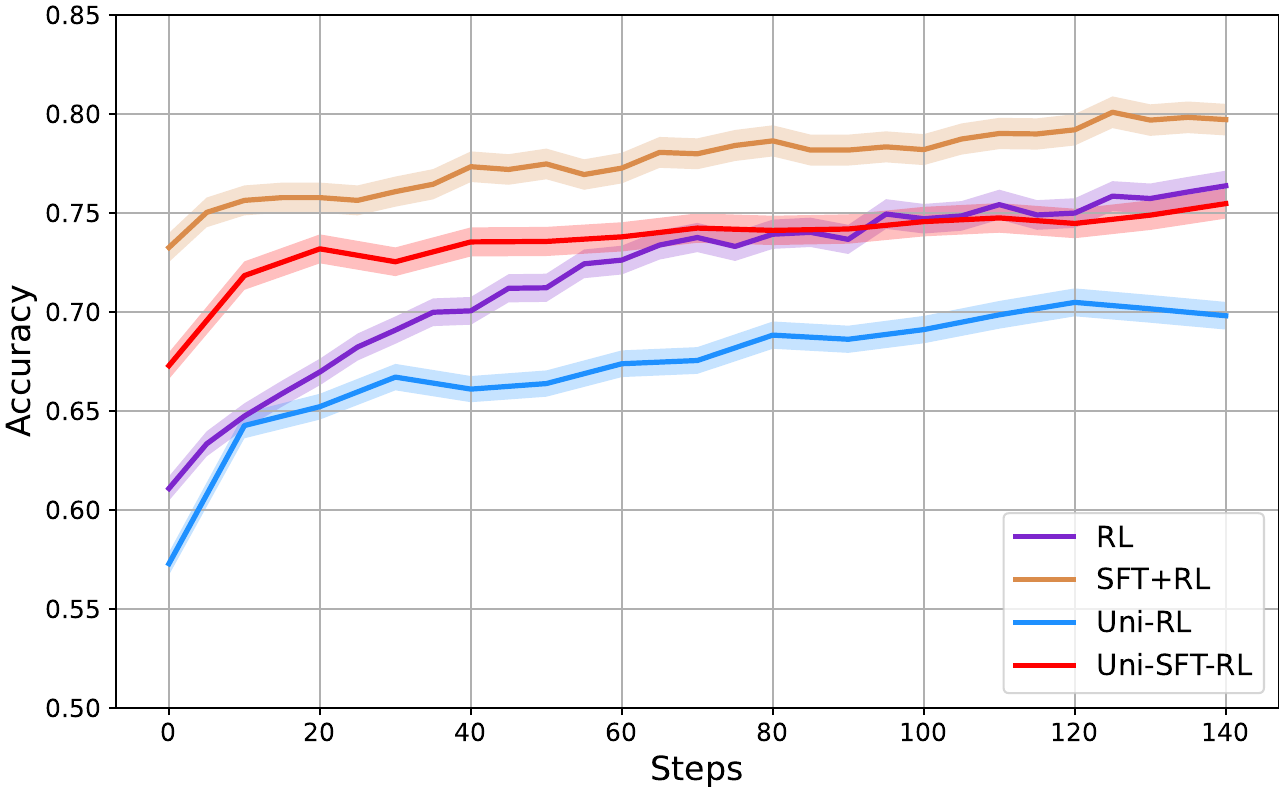}
        \caption{EM score with training steps.}
        \label{fig:wikitq_score_training}
    \end{subfigure}
    \caption{Training metrics for the WikiTQ dataset, showing the performance of RL and SFT+RL under both the single-dataset and unified-dataset settings.}
    \label{fig:training_metrics}
\end{figure}

\textbf{Training progress shows consistent improvement in reward and performance.}
In Figure \ref{fig:training_metrics}, we illustrate the trends of the reward score, model response length, and evaluation score over the training steps.
We observe that the response length of models with RL-Zero increases gradually with training steps. In contrast, the response length of models with SFT initially decreases and then stabilizes within a consistent range. Both reward and performance steadily improve with training steps.

\section{Analysis}
\label{sec:unified}

\subsection{Training Config Ablation Study}

\textbf{The format reward has a minor impact on the model's performance.}
From Table \ref{tab:reward_ablation} and Table \ref{tab:text_to_sql}, comparing 
$\lambda_1=0, \lambda_2=0$
 with 
$\lambda_1=0, \lambda_2=0.2$
, we observe that the format reward has a minimal impact on both text-to-SQL and overall experimental performance.

\textbf{RL enhances the text-to-SQL capabilities across different base models.} 
For text-to-SQL task, we observe that even with a strong base model, the Reasoning-Table approach still effectively improves performance.
When we apply RL using OmniSQL \cite{omnisql}—an SQL LLM trained on a high-quality SFT dataset comprising 200M data entries—its performance shows significant improvement compared to using Qwen-2.5-Coder \citep{hui2024qwen25coder}.
We also employed an additional technique by incorporating an extra reward based on N-gram similarity (App.\ref{sec:n_gram}). This involves matching the similarity between the predicted SQL and the gold SQL, which enables Reasoning-Table to nearly achieve the SOTA performance for 7B models on the BIRD dataset, achieving a performance of $68.3$ on the dev set.

\textbf{Position reward is beneficial for reinforcement learning in table reasoning.}
As shown in Table~\ref{tab:text_to_sql}, we conduct experiments using widely used datasets, Spider and BIRD. Comparing $\lambda_1=0, \lambda_2=0$ with $\lambda_1=0.2, \lambda_2=0$, we find that the position reward (similar to schema linking here) brings performance gains, improving the performance by $1.1\%$ on the BIRD dataset.
As shown in Table~\ref{tab:reward_ablation}, we analyze the impact of the position reward for three tableQA tasks. The configuration $\lambda_1=0, \lambda_2=0$ indicates that no position reward is used, while $\lambda_1=0.2, \lambda_2=0$ incorporates the position reward to ensure consistency. 
We observed that incorporating the position reward further improves the model's performance by approximately 0.6\% on WikiTQ and TAT-QA.
This suggests that the position reward helps the model better focus on the core content of the table and ensures consistency in its reasoning process and results. We believe that this type of reward still has room for further exploration to fully unlock its potential.

\textbf{LLM-as-a-judge can serve as a complement to rule-based string matching.}
Datasets like WikiTQ often contain multiple valid answers (e.g., ``2023-10-01'' vs. ``October 1, 2023'', or ``7 days'' vs. ``one week''), which rule-based evaluation methods, such as exact match (EM) or F1, struggle to handle due to their reliance on string matching. 
To address this limitation, we employ an LLM-as-a-judge approach, using an LLM \( M \) to assess the semantic equivalence or correctness of the predicted answer \( A \) relative to the reference answer \( A_{\text{ref}} \), producing a binary reward \( B_M(A, A_{\text{ref}}) \in \{0, 1\} \).
We use Qwen2.5-32B as the judge. The reward is defined as: $R_{\text{ans}} = B_M(A, A_{\text{ref}})$.
The reward \( B_M \) is computed by prompting \( M \) with the prompt \( P \), as detailed in Table~\ref{tab:llm_judge}. This approach effectively captures semantic equivalence (e.g., ``25°C'' vs. ``77°F''), overcoming the limitations of rule-based method.
In our experiments (Table~\ref{tab:reward_ablation}), we keep the reward weighting configuration constant (\( \lambda_1 = 0, \lambda_2 = 0 \)) and vary the method for obtaining \( R_{\text{ans}} \), comparing rule-based reward and LLM-judged reward, with prompts specified in Table~\ref{tab:llm_judge}. 
Results show that LLM-judged rewards improve performance on WikiTQ but have little effect on other datasets.

\begin{table}[htbp]
  \centering
  \begin{minipage}{0.48\textwidth}
    \centering
    \caption{Performance across different reward configurations, all using the \textit{Reason SFT+RL} setting.}
    \label{tab:reward_ablation}
    \resizebox{\textwidth}{!}{%
    \begin{tabular}{lccc}
      \toprule
      \textbf{Reward} & \textbf{WikiTQ} & \textbf{TAT-QA} & \textbf{HiTab} \\
      \midrule
      \multirow{2}{*}{\makecell[l]{$\lambda_1=0, \lambda_2=0$ \\ (Rule-based $R_{\text{ans}}$ )}} 
      & \multirow{2}{*}{80.35} & \multirow{2}{*}{79.62} & \multirow{2}{*}{81.38} \\ 
      & & & \\ \hdashline
      \multirow{2}{*}{\makecell[l]{$\lambda_1=0, \lambda_2=0.2$ \\ (Rule-based $R_{\text{ans}}$ )}} 
      & \multirow{2}{*}{80.07} & \multirow{2}{*}{79.79} & \multirow{2}{*}{81.94} \\ 
      & & & \\ \hdashline
      \multirow{2}{*}{\makecell[l]{$\lambda_1=0.2, \lambda_2=0$ \\ (Rule-based $R_{\text{ans}}$ )}} 
      & \multirow{2}{*}{80.95} & \multirow{2}{*}{79.47} & \multirow{2}{*}{81.60} \\
      & & & \\ 
      \hdashline
      \multirow{2}{*}{\makecell[l]{$\lambda_1=0, \lambda_2=0$ \\ (LLM-judged $R_{\text{ans}}$ )}} 
      & \multirow{2}{*}{81.23} & \multirow{2}{*}{79.96} & \multirow{2}{*}{81.04} \\
      & & & \\
      \bottomrule
    \end{tabular}
    }
  \end{minipage}
  \hfill
  \begin{minipage}{0.48\textwidth}
    \centering
    \caption{Performance across datasets with different models and reward configurations for text-to-SQL tasks.}
    \label{tab:text_to_sql}
    \resizebox{\textwidth}{!}{%
    \begin{tabular}{lcc}
      \toprule
      \textbf{Base Model + Reward} & \textbf{Spider} & \textbf{BIRD} \\
      \midrule
      \makecell[l]{Qwen2.5-Coder-Inst-7B \\ + Spider/BIRD train set SFT} & 82.9 & 59.5 \\ 
      \makecell[l]{+ $\lambda_1=0, \lambda_2=0$} & 83.3 & 62.1 \\ 
      \makecell[l]{+ $\lambda_1=0, \lambda_2=0.2$} & 83.4 & 62.7 \\
      \makecell[l]{+ $\lambda_1=0.2, \lambda_2=0$} & 84.2 & 63.1 \\ 
      \midrule
      \makecell[l]{OmniSQL-7B} & 84.8 & 64.1 \\  
      \makecell[l]{+ $\lambda_1=0, \lambda_2=0$} & 87.0 & 66.8 \\ 
      \makecell[l]{+ $\lambda_1=0.2, \lambda_2=0$} & 87.4 & 67.7 \\ 
      \makecell[l]{+ $\lambda_1=0.2, \lambda_2=0$ + N-gram} & 87.6 & 68.3 \\
      \bottomrule
    \end{tabular}
    }
  \end{minipage}
\end{table}

\subsection{Data Ablation Study}
\label{sec:data_ablation}
We conduct data ablation experiments in the RL-zero setting, based on $8$ sampling runs with a temperature of $0.8$, as outlined in Section 
 \ref{sec:rl_preprocess} and summarized in Table \ref{tab:rollout_datasets}.
We evaluate three data configurations.
\textit{All data}: the original, unfiltered dataset containing all examples;
\textit{Challenging data}: we filter out examples that were consistently answered correctly, retaining only those that presented challenges;
\textit{Variable-performance data}: we remove both examples that were consistently correct and those that were consistently incorrect, retaining only examples with variable performance.
We report the average performance across tableQA and text-to-SQL tasks for benchmarking, as shown in Figure \ref{fig:data_version}.
Notably, the model achieves the best performance when trained on \textit{Challenging data}.

\subsection{Advanced Evaluation and Analysis}

\begin{figure}[htbp]
  \centering
  \begin{minipage}[t]{0.45\textwidth}
    \centering
    \captionof{table}{Out-of-domain performance comparison between RL and SFT methods.}
    \label{tab:ood_tasks}
    \vspace{0pt} 
    \resizebox{\textwidth}{!}{%
      \begin{tabular}{lcc}
        \toprule
        \textbf{Model} & \textbf{AIT-QA} & \textbf{TableBench} \\
        \midrule
        Qwen2.5-Inst-7B & 86.21 & 51.22 \\
        Uni No-Reason SFT & 61.94 & 32.11 \\
        Uni Reason SFT & 62.33 & 51.80 \\
        Uni Zero-RL & 89.32 & 58.33 \\
        Uni Reason SFT+RL & \textbf{91.33} & \textbf{60.72} \\
        \bottomrule
      \end{tabular}%
    }
  \end{minipage}\hfill
  \begin{minipage}[t]{0.25\textwidth}
    \centering
    \captionof{figure}{Effect of different data selection strategy.}
    \label{fig:data_version}
    \vspace{0pt} 
    \includegraphics[width=\textwidth]{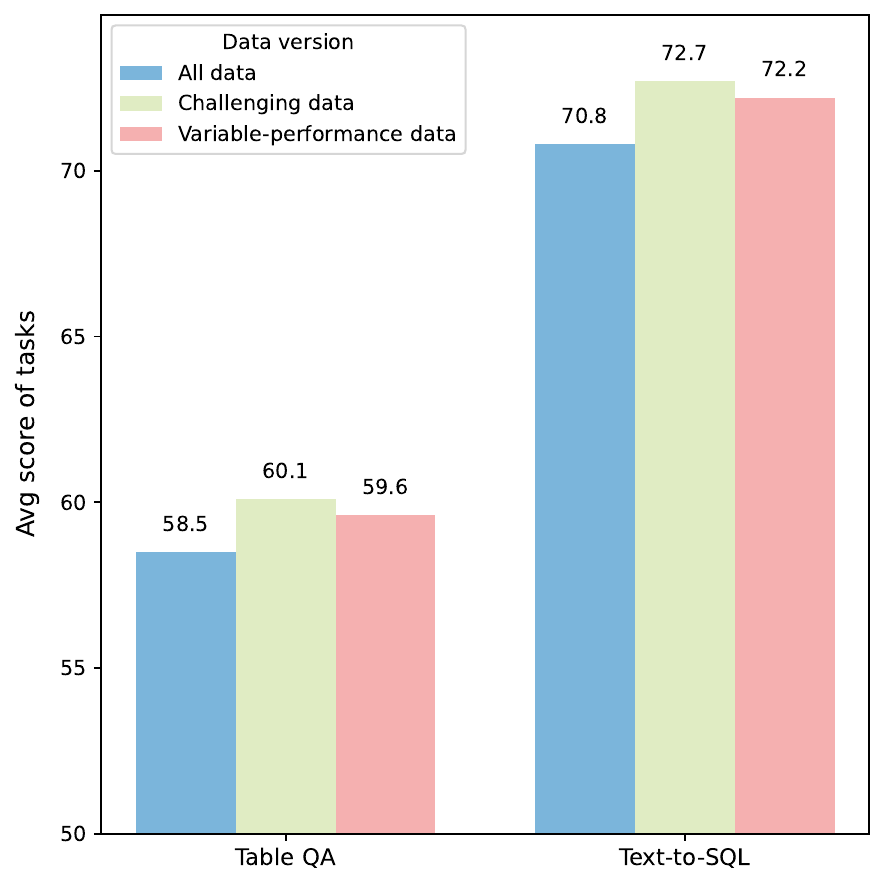}
  \end{minipage}\hfill
  \begin{minipage}[t]{0.23\textwidth}
    \centering
    \captionof{figure}{Impact of model size on performance (Avg. tableQA).}
    \label{fig:score_trend}
    \vspace{0pt} 
    \includegraphics[width=\textwidth]{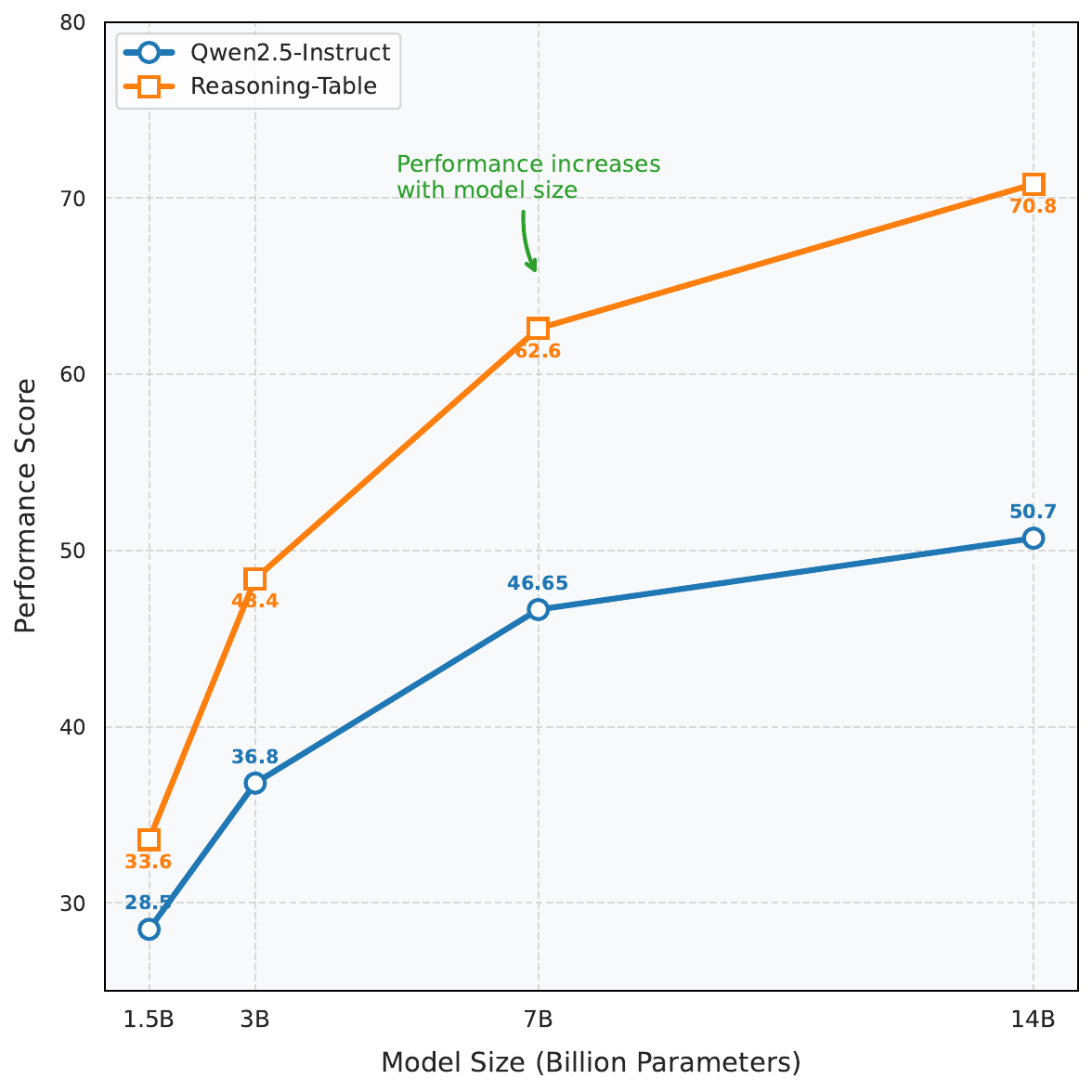}
  \end{minipage}
\end{figure}

\textbf{Robustness evaluation.}
Reasoning-Table enhances the model's ability to handle table perturbations, including (1) \textit{input format perturbation} and (2) \textit{column order perturbation}. 
As shown in Table \ref{tab:table_formats},
we use different table serialization methods as input (markdown was used during training).
RL models exhibit robust and superior performance under various input format perturbations.
We keep the question unchanged while randomly altering the row and column order of the table.
According to Table \ref{tab:table_perturbations}, SFT models are highly sensitive to column and row perturbations, exhibiting a performance drop of $15.22\%$. In contrast, RL models adapt effectively to these changes, as they develop strong table reasoning skills through reinforcement learning, rather than relying solely on imitation.

\begin{table}[htbp]
  \centering
  \begin{minipage}[t]{0.48\textwidth}
    \centering
    \caption{Different input formats of WikiTQ Task. We using the single dataset training setting.}
    \label{tab:table_formats}
    \vspace{0pt} 
    \resizebox{\textwidth}{!}{%
      \begin{tabular}{lccc}
        \toprule
        \textbf{Model} & \textbf{Markdown} & \textbf{CSV} & \textbf{DataFrame} \\
        \midrule
        Qwen2.5-Inst-7B & 57.27 & 59.23 & 58.72 \\
        No-Reason SFT & 58.19 & 56.97 & 55.29 \\
        Reason SFT & 74.88 & 69.44 & 70.26 \\
        Zero-RL & 76.35 & 76.35 & 72.26 \\
        Reason SFT+RL & 80.07 & 80.45 & 78.95 \\
        \bottomrule
      \end{tabular}%
    }
  \end{minipage}\hfill
  \begin{minipage}[t]{0.48\textwidth}
    \centering
    \caption{Model performance under perturbations, with \% drops from original setting. Higher scores, lower drops show better robustness.}
    \label{tab:table_perturbations}
    \vspace{0pt} 
    \resizebox{\textwidth}{!}{%
    \begin{tabular}{l|cccc}
        \toprule
        \textbf{Model} & \textbf{Origin} & \textbf{Column} & \textbf{Row} & \textbf{Both} \\
        \midrule
        Qwen2.5-Inst-7B & 57.27 & 56.80\textcolor{gray}{\tiny{(-0.47)}} & 51.56\textcolor{gray}{\tiny{(-5.71)}} & 47.26\textcolor{gray}{\tiny{(-10.01)}} \\
        \midrule
        Uni No-Reason SFT & 57.21 & 53.61\textcolor{gray}{\tiny{(-3.60)}} & 48.82\textcolor{gray}{\tiny{(-8.39)}} & 46.38\textcolor{gray}{\tiny{(-10.83)}} \\
        Uni Reason SFT & 72.35 & 68.46\textcolor{gray}{\tiny{(-3.89)}} & 64.70\textcolor{gray}{\tiny{(-7.65)}} & 57.66\textcolor{gray}{\tiny{(-14.69)}} \\
        Uni Zero-RL & 72.75 & 70.31\textcolor{gray}{\tiny{(-2.44)}} & 68.58\textcolor{gray}{\tiny{(-4.17)}} & 66.80\textcolor{gray}{\tiny{(-5.95)}} \\
        Uni SFT+RL & \textbf{75.46} & \textbf{74.50}\textcolor{gray}{\tiny{(-0.96)}} & \textbf{70.80}\textcolor{gray}{\tiny{(-4.66)}} & \textbf{68.20}\textcolor{gray}{\tiny{(-7.26)}} \\
        \bottomrule
    \end{tabular}%
    }
  \end{minipage}
\end{table}

\textbf{Generalization assessment.}
We evaluate the generalization of Reasoning-Table on out-of-domain datasets, featuring task and table types not encountered during training. Specifically, we use AIT-QA~\citep{aitqa} and TableBench~\citep{wu2025tablebench} (see App.\ref{app:Out-of-domain} for details), with results presented in Table~\ref{tab:ood_tasks}.
The results show that the RL method outperforms the SFT method by approximately $30\%$ on unseen datasets. 
This indicates that SFT's imitation learning often limits the model's generalization ability, thereby impairing its general reasoning capabilities. 
In contrast, Reasoning-Table, by optimizing rewards through exploration, significantly enhances the model's adaptability and robustness across diverse table reasoning tasks.

\textbf{Scaling training analysis.}
The scaling law, which describes how model size impacts performance, is essential for evaluating the efficacy of training approaches. We examine the scaling properties of the proposed training method by testing models of varying sizes from the Qwen2.5-Instruct series, specifically 1.5B, 3B, 7B, and 14B parameters. We measure average performance on the tableQA benchmarks, with comprehensive results displayed in Figure \ref{fig:score_trend}. Larger models consistently achieve better results than smaller ones across both tasks. However, post-training performance gains are minimal for smaller models (1.5B), while larger models show significant improvements. These results indicate that the R1-paradigm training approach becomes increasingly effective as model size grows.

\begin{wrapfigure}{t}{0.6\textwidth}
    \centering
    \includegraphics[width=0.47\textwidth]{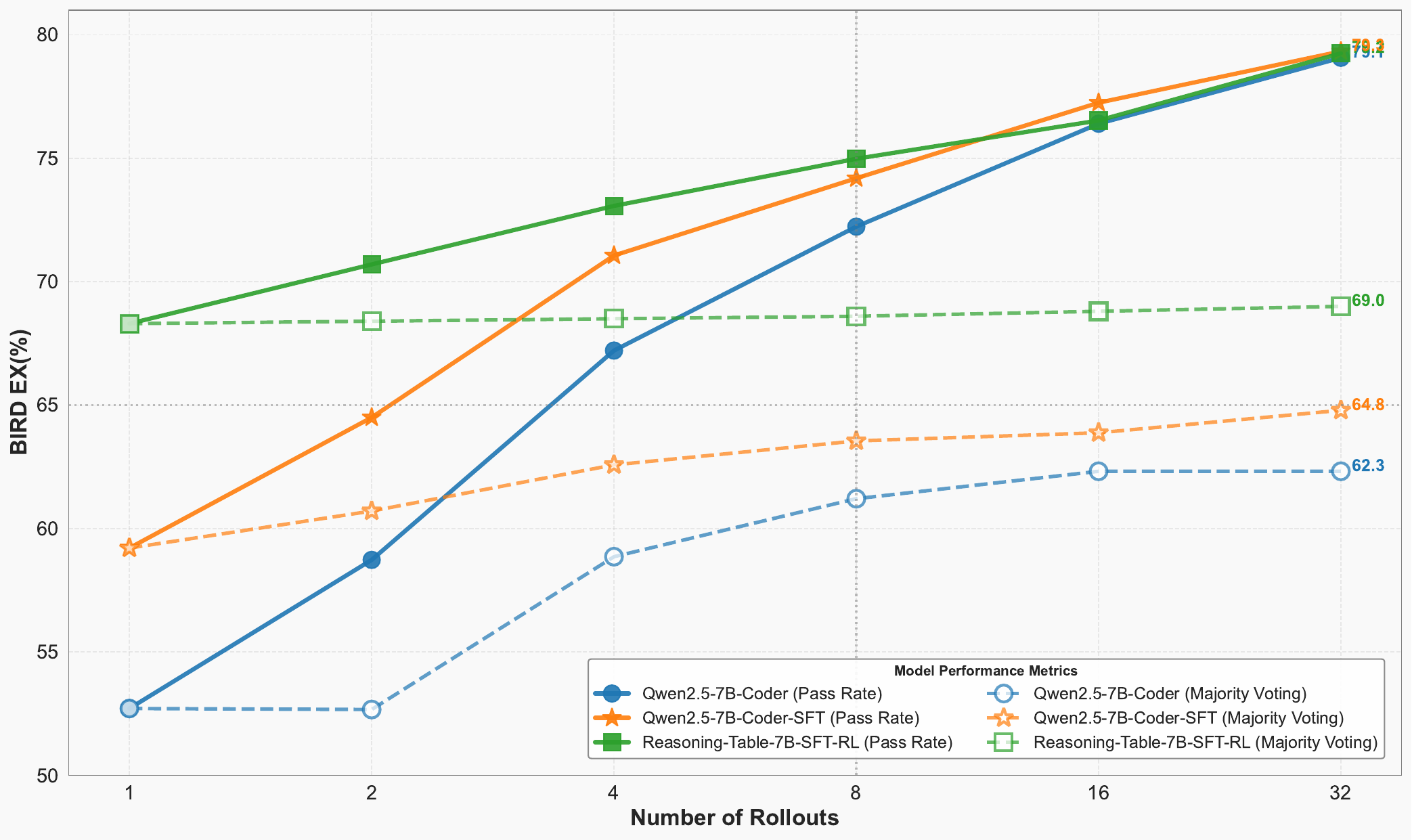}
    \caption{Test-Time Scaling. For the BIRD dataset, we perform multiple samplings and report the pass@n and majority voting EX scores.}
    \label{fig:text_to_sql_voting}
\end{wrapfigure}

\subsection{Test time scaling analysis.}
As shown in Figure~\ref{fig:text_to_sql_voting}, we evaluate the impact of RL training on test-time scaling in the text-to-SQL task, using Qwen2.5-7B-Coder, Qwen2.5-7B-Coder-SFT, and Reasoning-Table-7B-SFT-RL, with rollouts of 1, 2, 4, 8, 16, and 32, respectively. The line graph illustrates pass rate and majority voting performance across these rollouts, demonstrating that RL training significantly enhances the model’s test-time scaling capability. Using the strongest 
OmniSQL model as the base, Reasoning-Table achieves a performance of $69.0$ in the majority voting setting, approaching single-model state-of-the-art results on the BIRD-Bench leaderboard (\url{https://bird-bench.github.io}).

\vspace{-5pt}
\section{Related Work}
\label{sec:related}

\textbf{Table reasoning.}
Table reasoning focuses on enabling models to understand and reason over structured tabular data to perform tasks such as question answering, fact verification, and text generation. Several datasets have been proposed, including those for table question answering \cite{wikisql,wikitablequestion,zhu2021tat,chen2020hybridqa,chen2021finqa}, fact verification \citep{tabfact,aly2021feverous}, table-to-text \citep{parikh2020totto}, and text-to-SQL \cite{yu2018spider,li2024bird}, providing a wide range of datasets for the field. There are two main approaches: one involves carefully designed prompt engineering, leveraging large language models combined with tool calls like Python, SQL to address tasks \citep{cheng2022binder,api-table,jiang-etal-2023-structgpt,chain-of-table,ye2023table,cao2025tablemaster}. The other approach uses fine-tuning with table-specific data to enable models to understand tables \cite{xie2022unifiedskg, zhang-etal-2024-tablellama,zhang2024tablellm,su2024tablegpt2,zhu2024tatllm}.
\ours not only uses large amounts of table data for training but also incorporates RL training, making it the first work to experiment and analyze across all table reasoning tasks.
\ours shares the same motivation as Table-R1 \citep{yang2025tabler1_yale, wu2025tabler1_beihang}, but our work differs in data and experimental settings. We believe that these studies can jointly contribute to advancing table reasoning using reinforcement learning.

\textbf{RL for reasoning.}
The emergence of DeepSeek-R1 \citep{guo2025deepseek} shows RL's ability to guide models in generating high-quality reasoning trajectories. This has led to studies enhancing reasoning in mathematics (e.g., DeepScaleR~\citep{deepscaler2025}, AReaL~\citep{mei2025real}, Light-R1~\citep{wen2025light}, DAPO~\citep{yu2025dapo}), VAPO~\citep{yuan2025vapo}, and code generation (e.g., DeepCoder~\citep{deepcoder2025}), with significant progress.
The R1-paradigm has made breakthroughs in various NLP domains, such as tool use~\citep{feng2025retool,qian2025toolrl,zhang2025nemotron}, RAG~\citep{jin2025search}, and SQL generation~\citep{reasoningsql,sqlr1}. Our work focuses on table reasoning, conducting comprehensive training and analysis.

\vspace{-5pt}
\section{Conclusion}
\label{sec:conclusion}

We propose \ours, the first work to apply the LLM reinforcement learning with the rule-based outcome reward for training on table reasoning tasks, covering a diverse range of task types and tables. 
Experiments on tasks such as tableQA and text-to-SQL demonstrate that \ours significantly enhances table reasoning capabilities, achieving performance comparable to proprietary LLMs. 
Additionally, \ours improves model robustness and generalization. 
It also provides high-quality reasoning datasets, detailed table-processing solutions, data selection strategies, and evaluation scripts, laying a foundation for future scaling of table-reasoning training.

\section*{Limitations}

\ours is the first work to apply reinforcement learning to table reasoning tasks for large language models. We identify two limitations in our work, which also point to our future directions:

\textbf{Lack of a code execution module.} While we used an SQL executor for text-to-SQL tasks, we did not integrate a code execution module for other tasks such as tableQA, which can introduce potential hallucinations and computational errors. In the future, we will explore table RL based on code execution and tool integration.

\textbf{Lack of fine-grained control over data proportions.}
As shown in Table \ref{tab:tqa_result}, when we perform combined training using all table QA datasets, the model's performance is worse than that of single-dataset training. This is caused by differences in the distribution of training data and rewards. We believe that more effective data combination strategies could help alleviate this issue. Looking ahead, exploring how to achieve truly unified training across all table reasoning tasks will be an important research direction.

\textbf{Lack of data scaling across diverse sources.} Although we utilized nearly all available table reasoning datasets for unified training, this approach remains somewhat in-domain and lacks scalability. In the future, we will investigate a pipeline for automatically collecting and synthesizing table reasoning data from the Internet to enable scalable RL training for table reasoning.

\bibliography{colm2025_conference}
\bibliographystyle{colm2025_conference}

\appendix

\appendix
\onecolumn
\clearpage

\section{Data preprocess details}

\subsection{Reason-SFT Data Quailty Control.}
\label{app:reason_sft_details}
Since reasoning process generated by Deepseek-R1 may include redundant or contradictory steps (e.g., the model revises earlier correct reasoning unnecessarily), we introduce a redundancy detection algorithm to eliminate such low-quality samples. Specifically, we focus on the text within \texttt{<think>} tags and apply the following steps:\\
\textit{(1) Sentence Segmentation and Filtering}: Each reasoning segment is split into individual sentences. Sentences with fewer than five words are discarded to remove trivial or malformed content.\\
\textit{(2) TF-IDF and Cosine Similarity Computation}: Sentences are vectorized using TF-IDF, and pairwise cosine similarity scores are computed for all sentence pairs within each \texttt{<think>} segment.\\
\textit{(3)Semantic Similarity Penalty Adjustments}:
If one sentence is a question and the other is not, the similarity score is down-weighted (Question Mismatch Penalty).
If two similar sentences differ significantly in modal certainty (e.g., use of \textit{might}, \textit{must}, \textit{will}), their similarity score is penalized (Modal Verb Disparity Penalty).
The final adjusted similarity score between sentence pair $(s_i, s_j)$ is computed as:
\[
\begin{aligned}
\tilde{sim}(s_i, s_j) &= sim_{\text{cos}}(s_i, s_j) \cdot P_{\text{QM}}(s_i, s_j) \cdot P_{\text{MV}}(s_i, s_j)
\end{aligned}
\]
where $P_{\text{QM}} \in [0,1]$ and $P_{\text{MV}} \in [0,1]$ are penalty factors for question mismatch and modal verb disparity, respectively.
\textit{(4) Redundancy-Based Filtering Criterion}:  
If more than two sentence pairs within a \texttt{<think>} block exhibit high adjusted similarity (threshold $\tau = 0.7$), the entire example is marked as redundant and removed. Formally, let $S = \{s_1, s_2, \dots, s_n\}$ be the set of valid sentences in a \texttt{<think>} segment. Define:
\[
    R = \left\{ (s_i, s_j) \mid \tilde{sim}(s_i, s_j) > \tau,\ i < j \right\}
\]
If $|R| > 2$, the entire example is marked as redundant and removed.

\subsection{SFT Data Quailty Control}
\label{sec:sft_data_quailty}

The data processing pipeline shown in Table~\ref{tab:filtered_datasets} presents statistics on the number of samples retained during SFT data collection.

\begin{table}[htbp]
  \centering
  \caption{Filtered table reasoning datasets. The Origin column represents the amount of No-Reason Data;; After Stage1 is the amount of Reason-SFT data synthesized by RFT on the training set; After Stage2 is the amount of Reason-SFT data finally retained after our filtering algorithm}
  \label{tab:filtered_datasets}
  \resizebox{0.6\textwidth}{!}{%
    \begin{tabular}{lcccc}
      \toprule
      \textbf{Dataset} & \textbf{\makecell{Origin \\ (No-Reason Data)}} & \textbf{After Stage1} & \textbf{\makecell{After Stage2 \\ (Reason-SFT Data)}} & \textbf{\makecell{Unified \\ SFT Data}} \\
      \midrule
      WikiTQ & 14,152 & 11,306 & 10,446 & 10,446 \\
      HybridQA & 62,682 & 41,734 & 38,544 & 5,000 \\
      MultiHiertt & 7,830 & 1,879 & 1,475 & 1,475 \\
      OTT-QA & 41,469/10,000 & 6,031 & 5,357 & 5,357 \\
      FinQA & 6,251 & 3,774 & 3,321 & 3,321 \\
      FeTaQA & 7,326 & 2,819 & 2,663 & 2,663 \\
      TAT-QA & 13,210 & 8,553 & 7,559 & 7,559 \\
      HiTab & 7,417 & 4,887 & 4,380 & 4,380 \\
      ToTTo & 120,761/6,500 & 2,097 & 1,896 & 0 \\
      TabFact & 92,283/24,101 & 21,238 & 19,029 & 3,000 \\
      FEVEROUS & 71,291/4,000 & 3,024 & 2,894 & 0 \\
      \hline
      Total & 444,672 & 107,342 & 97,564 & 43,201 \\
      \bottomrule
    \end{tabular}%
  }
\end{table}

\subsection{RL Data Quailty Control}
\label{sec:rl_data_quailty}

\textbf{Difficulty control.}
We use Qwen2.5-32B-Instruct to answer each question in the training set 8 times to classify the difficulty of each example. This was done only for the tableQA and text2sql tasks. The experimental results are shown in Table \ref{tab:rollout_datasets}.\\
\textbf{Position evidence control.}
We extract position evidence for three datasets, WikiTQ, TAT-QA and HiTab, the statistic results are shown in Table \ref{tab:performance_datasets}.\\

\begin{table}[h]
  \centering
  \caption{Table reasoning datasets with \textit{Pass@n} performance.}
  \label{tab:rollout_datasets}
  \resizebox{0.48\textwidth}{!}{%
    \begin{tabular}{lcccc}
      \toprule
      \textbf{Dataset} & \textbf{Pass@1} & \textbf{Pass@2} & \textbf{Pass@4} & \textbf{Pass@8} \\
      \midrule
      WikiTQ & 76.80  & 77.03 & 78.89  & 79.72 \\
      HybridQA & 38.26 & 40.80 & 42.36 & 43.72 \\
      MultiHiertt & 30.54 & 34.61 & 36.56 & 39.15 \\
      OTT-QA & 31.26 & 34.03 & 37.14 & 39.24 \\
      FinQA & 61.29 & 66.17 & 68.96 & 71.14 \\
      TAT-QA & 57.94 & 64.95 & 69.44 & 73.76 \\
      HiTab & 73.29 & 77.15 & 80.18 & 81.94 \\
      Spider & 83.36 & 88.29 & 89.65 & 91.49 \\
      BIRD & 62.19 & 69.17 & 72.29 & 75.10 \\
      \bottomrule
    \end{tabular}%
  }
\end{table} 


\begin{table}[htbp]
  \centering
  \caption{Position evidence generation control. Solve Rate indicates that the model predicts the correct answer, we use Pass@10; Evidence-Containing Responses indicates the number of responses that contain evidence in the reasoning process; Valid Evidence Responses indicates that the evidence in the reasoning process is true from the table; Overall Correct denotes the accuracy of both correct answers and evidence from the table.}
  \label{tab:performance_datasets}
  \resizebox{0.95\textwidth}{!}{%
    \begin{tabular}{lccccc}
      \toprule
      \textbf{Dataset} & \textbf{Solve Rate (Pass@10)} & \textbf{Evidence-Containing Responses (\%)}  & \textbf{Valid Evidence Responses (\%)} & \textbf{Overall Correct (\%)} \\
      \midrule
      WikiTQ & 86.66 & 95.80 &  98.59 & 79.35 \\
      TAT-QA & 74.00 & 89.95 & 72.57 & 67.29 \\
      HiTab & 71.35 & 99.58 & 99.91 & 70.58 \\ 
     \midrule
      Average & 78.53 & 94.53 & 91.28 & 71.17 \\
      \bottomrule
    \end{tabular}%
  }
\end{table}

\subsection{Example of Filtered Data}

Our filtering strategy effectively removes redundant reasoning processes. In Table \ref{tab:badcase}, we present an example that was filtered out due to excessive repetitive verification and ineffective reasoning.

\begin{longtable}{p{0.95\textwidth}}
\caption{A filtered case. The reasoning process is wrapped in \think{}, and one can clearly see that this reasoning process is redundant.} \\
\vspace{-1.5em} 
\label{tab:badcase} \\
\hline
\textbf{Question}: How many people entered office in 1970? \\
\hline
\textbf{Ground Truth}: 10 \\
\hline
\endfirsthead

\hline
\endhead

\hline
\endfoot

\hline
\endlastfoot

\think{
Alright, let's tackle this question. The task is to find out how many people entered office in 1970 based on the provided table. Let me start by understanding the table structure and then check each entry carefully. \\

First, the table has four columns: \textbf{State, Title, Name, and Entered office}. The "Entered office" column has years, some with exact dates and others just the year. \textcolor{red}{The question specifically asks for entries in 1970. So, I need to scan through the "Entered office" column and count all instances where the year is 1970.} \\

Let me go row by row. \\
1. Barbados - Sir Stanley Robinson entered in 1966. Not 1970. \\
2. Barbados - Sir Theodore Brancker entered in 1961. Nope. \\
3. Bolivia - President of the Senate, entered 1970. Name is missing, but the year is 1970. So that's one. \\

\ldots \\

\textcolor{red}{Wait, let me recount:} \\
- Bolivia: 2 entries (President of the Senate and Chamber of Deputies) both 1970. Total 2. \\
- Brazil: 2 entries (President of the Senate and Chamber of Deputies) both 1970. Now total 4. \\
- El Salvador: 1970 (with ?). Total 8. \\
- Guatemala: 1970. Total 9. \\
- Uruguay: President of the Chamber of Deputies entered March 1, 1970. Total 10. \\

\textcolor{red}{Wait, that's 10? Let me check again.} \\
Wait, the entries are: \\
1. Bolivia President of the Senate: 1970 (1) \\
2. Bolivia President of the Chamber of Deputies: 1970 (2) \\
10. Uruguay President of the Chamber of Deputies: March 1, 1970 (10) \\

Additionally, Nicaragua has entries with "None" so excluded. But the entry for Uruguay's Fernando Elichirigoity is March 1, 1970, which counts. \\

Wait, also, what about Venezuela? Their entries are 1968. No. Mexico's entries are 1967. So the count is 10? \\

\textcolor{red}{Wait, but let me check the exact entries again:} \\
Looking at the table: \\
- Bolivia: two entries with 1970. So two. \\
- Brazil: two entries 1970. So two more, total 4. \\
- Uruguay: Chamber of Deputies entered March 1, 1970. 10. \\

\textcolor{red}{Is there any other entry? Let's check again:} \\
Looking at the list: \\
- Ecuador: both entries have ?. So no. \\
- Puerto Rico: 1969. No. \\
- Others are all prior years. \\

So total of 10 entries. But \textcolor{red}{wait, let me check again the entries:} \\
Wait, the Uruguay entry for Fernando Elichirigoity is March 1, 1970. So that's definitely 1970. So yes, that's 10. \\

\textcolor{red}{But wait, the El Salvador entry has a question mark. Should that be counted?} \\

The original question says "how many people entered office in 1970" based on the table. The table lists some entries with a question mark. For example, El Salvador's entry is "1970 ?". The question mark here might indicate uncertainty in the data. However, the problem statement says to answer based on the table as presented, so even if there's a question mark, if the year is written as 1970, we should count it. Similarly, entries like "1966 ?" (Dominican Republic) would not be counted as 1970. \\

Therefore, El Salvador's entry with "1970 ?" is considered as 1970. So that's included, making the total 10. \\

\textcolor{red}{Wait, but let's recount:} \\
1. Bolivia President of the Senate: 1970 (1) \\
4. Brazil President of the Chamber of Deputies: 1970 (4) \\
10. Uruguay President of the Chamber of Deputies: 1970 (10) \\

Yes, that's 10. Therefore, the answer is 10. \\

\textcolor{red}{But wait, I need to check if any of the entries with exact dates (e.g., July 20, 1970) are counted. Since the year is 1970, they should be counted regardless of the month. So even if they entered in July, it's still 1970. So those are valid.} \\

Now, cross-checking again, the entries are: \\
- Bolivia: 2 \\
- Brazil: 2 \\
- Chile: 1 \\
- Uruguay: 1 \\

Total: $2 + 2 = 4$, plus $1 = 5$, plus $1 = 6$, plus $1 = 7$, plus $1 = 8$, plus $1 = 9$, plus $1 = 10$. \\

Yes, that's 10. So the answer should be 10.} \\
\hline
\answer{10}\\
\hline
\end{longtable}

\section{Evaluation Details}
\label{sec:appendix}

\subsection{Datasets details}
\label{sec:datasets_and_metrics}
For OTT-QA \citep{ottqa}, which is an open retrieval QA dataset, we use the oracle setting. This means that we only use the provided table containing the correct answer as input. The focus of our paper is on table reasoning, rather than retrieval.
For HybridQA \citep{chen2020hybridqa} and MultiHiertt \citep{zhao2022multihiertt}, we use the data processed by TableQAKit \citep{tableqakit}. They performed preprocessing, filtering out irrelevant long passages and retaining only the tables and some core text, which is beneficial for our study of table reasoning.

\subsection{Tasks and Metrics}
\label{sec:tasks_and_metrics}
The tasks and metrics are shown in Table \ref{tab:datasets}. 
\begin{table*}[h]
  \centering
  \caption{All table reasoning tasks. Tasks marked with \textsuperscript{*} indicate out-of-domain evaluation.}
  \label{tab:datasets}
  \resizebox{0.9\textwidth}{!}{%
    \begin{tabular}{llccc}
      \toprule
      \textbf{Task Type} & \textbf{Task} & \textbf{Table Type} & \textbf{Output} & \textbf{Metrics} \\
      \midrule
      \multirow{8}{*}{Question Answering} & WikiTQ~\citep{pasupat2015compositional} & Simple Table & Answer & Acc\\
      & HybridQA~\citep{chen2020hybridqa} & Semi-Structured Table & Answer & EM,F1\\
      & MultiHiertt~\citep{zhao2022multihiertt} & Spreadsheet Table & Answer & EM,F1\\
      & OTT-QA~\citep{ottqa} & Semi-Structured Table & Answer & EM,F1 \\
      & FinQA~\citep{chen2021finqa} & Spreadsheet Table & Answer & Acc \\
      & FeTaQA~\citep{nan2022fetaqa} & Spreadsheet Table & Sentence Answer & BLEU\\
      & TAT-QA~\citep{nan2022fetaqa} & Spreadsheet Table & Free-Form Answer & EM,F1 \\
      & HiTab~\citep{cheng2022hitab} & Spreadsheet Table & Answer & Acc \\
      & AIT-QA\textsuperscript{*}~\citep{nan2022fetaqa} & Simple Table & Answer & EM \\
      & TableBench\textsuperscript{*}~\citep{wu2025tablebench} & Simple Table & Answer & EM \\
      \midrule
      \multirow{2}{*}{Text-to-SQL} & Spider \cite{yu2018spider} & Database Table & SQL & EX \\
      & BIRD \cite{li2024bird} & Database Table & SQL & EX \\
      \midrule
      \multirow{1}{*}{Table-to-Text} & ToTTo~\citep{parikh2020totto} & Simple Table & Text & BLEU\\
      \midrule
      \multirow{2}{*}{Fact Verification} & TabFact~\citep{tabfact} & Simple Table & Boolean & Acc\\
      & FEVEROUS~\citep{aly2021feverous} & Simple Table & Boolean & Acc\\
      \bottomrule
    \end{tabular}%
  }
\end{table*}

\subsection{Out-of-domain Evaluation}
\label{app:Out-of-domain}

We evaluate out-of-domain generalization on TableBench~\citep{wu2025tablebench} and AIT-QA~\citep{aitqa}. For TableBench, we use two subsets: \textsc{NumericalReasoning} and \textsc{FactChecking}, with exact match (EM) as the evaluation metric. For AIT-QA, we report execution accuracy (EX). All rollouts are generated using vLLM~v0.8.2 with a maximum generation length of 4096 tokens.

\subsection{Metrics}
\label{app:metrics}

\textbf{Exact Match (EM).} This metric returns 1 if the predicted answer exactly matches the reference answer, and 0 otherwise:
\[
\text{EM} = 
\begin{cases}
1 & \text{if } \hat{y} = y \\
0 & \text{otherwise}
\end{cases}
\]
where $\hat{y}$ is the predicted answer and $y$ is the ground-truth answer.\\
\textbf{BLEU Score.} BLEU measures the n-gram precision with a brevity penalty (BP) to penalize short candidates:
\[
\text{BLEU} = \text{BP} \cdot \exp\left(\sum_{n=1}^{N} w_n \log p_n\right)
\]
where $p_n$ is the modified n-gram precision, $w_n$ is the weight (typically uniform), and $\text{BP}$ is defined as:
\[
\text{BP} = 
\begin{cases}
1 & \text{if } c > r \\
\exp\left(1 - \frac{r}{c}\right) & \text{if } c \leq r
\end{cases}
\]
with $c$ the length of the candidate and $r$ the length of the reference.\\
\textbf{Execution Accuracy (EX).} This metric evaluates whether the execution result of the predicted logical form equals the ground-truth execution result:
\[
\text{EX} = 
\begin{cases}
1 & \text{if } \text{Execute}(\hat{P}) = \text{Execute}(P) \\
0 & \text{otherwise}
\end{cases}
\]
where $\hat{P}$ and $P$ are the predicted and gold programs/logical forms respectively, and \texttt{Execute} denotes the result of executing the program on the input table.\\
\textbf{F1 Score.}
The F1 score is computed based on the precision and recall of tokens in \( A \) relative to \( A_{\text{ref}} \), where precision is the fraction of tokens in \( A \) that appear in \( A_{\text{ref}} \), and recall is the fraction of tokens in \( A_{\text{ref}} \) that appear in \( A \).
\[
\text{Precision} = \frac{|\text{Tokens}(A) \cap \text{Tokens}(A_{\text{ref}})|}{|\text{Tokens}(A)|}, \quad \text{Recall} = \frac{|\text{Tokens}(A) \cap \text{Tokens}(A_{\text{ref}})|}{|\text{Tokens}(A_{\text{ref}})|}
\]
\[
\text{F1} = 2 \cdot \frac{\text{Precision} \cdot \text{Recall}}{\text{Precision} + \text{Recall}}
\]
Tokens are obtained by splitting \( A \) and \( A_{\text{ref}} \) into words or subwords, ignoring case and punctuation, unless specified otherwise.\\

\section{Experiments Details}
\label{sec:exp_details}

\subsection{Setup}

\textbf{SFT settings.}
For single-dataset SFT, We perform full fine-tuning separately on the no-reason and reason data for each task. we use a batch size of 64, a learning rate of 5e-6, and train for 5 epochs.

For uni-dataset SFT, We merge the no-reasoning and reasoning data from all tasks in the Table \ref{tab:datasets} and perform full fine-tuning on each. We use a batch size of 64, a learning rate of 5e-6, and train for 5 epochs.

\textbf{RL setting.}
We adopt the GRPO algorithm as our core reinforcement learning method and enhance it with several modifications. We discard the entropy loss entirely. To prevent rapid entropy collapse during training, we follow DAPO~\cite{yu2025dapo} and apply the Clip-Higher strategy by setting the clip ratio upper bound to 0.28. To further reduce the noise introduced by excessively long sequences, we mask the loss of truncated responses. The rollout temperature is fixed at 1.0 for all experiments, and the learning rate is set to \( 1 \times 10^{-6} \). 

For SQL RL, we remove the KL divergence loss term. We use a rollout batch size of 64 and an update batch size of 16. For each prompt, we sample 16 trajectories during rollout. The maximum response length is set to 8192 tokens.

For single-dataset RL and uni-dataset RL, we retain the KL loss with a small coefficient of 0.001. We use a rollout batch size of 256 and an update batch size of 64. Each prompt is sampled with 5 trajectories during rollout. The maximum response length is set to 4096 tokens.

\subsection{N-Gram similarity for SQL.}
\label{sec:n_gram}
For predicted SQL \( S_p \) and gold SQL \( S_g \):
\begin{enumerate}
    \item \textbf{Tokenize}: Split into \( T_p = \{t_{p,1}, \dots, t_{p,m}\} \), \( T_g = \{t_{g,1}, \dots, t_{g,n}\} \), normalizing keywords.
    \item \textbf{N-grams}: For \( n \):
    \[
    N_p = \{ (t_{p,i}, \dots, t_{p,i+n-1}) \mid 1 \leq i \leq m - n + 1 \}
    \]
    \[
    N_g = \{ (t_{g,j}, \dots, t_{g,j+n-1}) \mid 1 \leq j \leq n - n + 1 \}
    \]
    \item \textbf{Similarity}:
    \[
    \text{sim}_n = \begin{cases} 
        \frac{|N_p \cap N_g|}{|N_p \cup N_g|} & \text{if } |N_p \cup N_g| > 0 \\
        0 & \text{otherwise}
    \end{cases}
    \]
\end{enumerate}
Use \( n=2 \); if tokens are insufficient, set \( n=1 \).



\section{Other train results}
\begin{figure*}[h]
    \centering
    \begin{subfigure}[t]{0.32\textwidth}
        \centering
        \includegraphics[width=\textwidth]{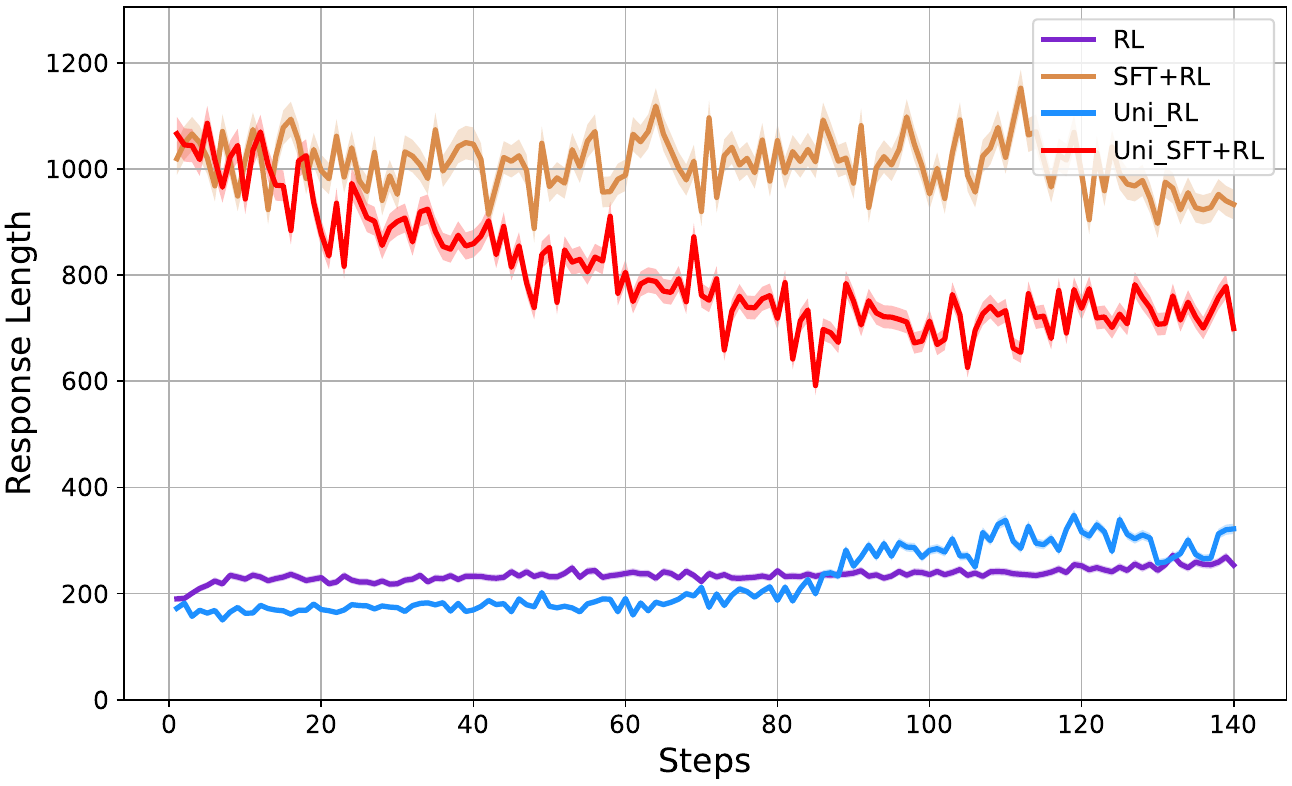}
        \caption{Response length with training steps.}
        \label{fig:length_training}
    \end{subfigure}
    \hfill
    \begin{subfigure}[t]{0.32\textwidth}
        \centering
        \includegraphics[width=\textwidth]{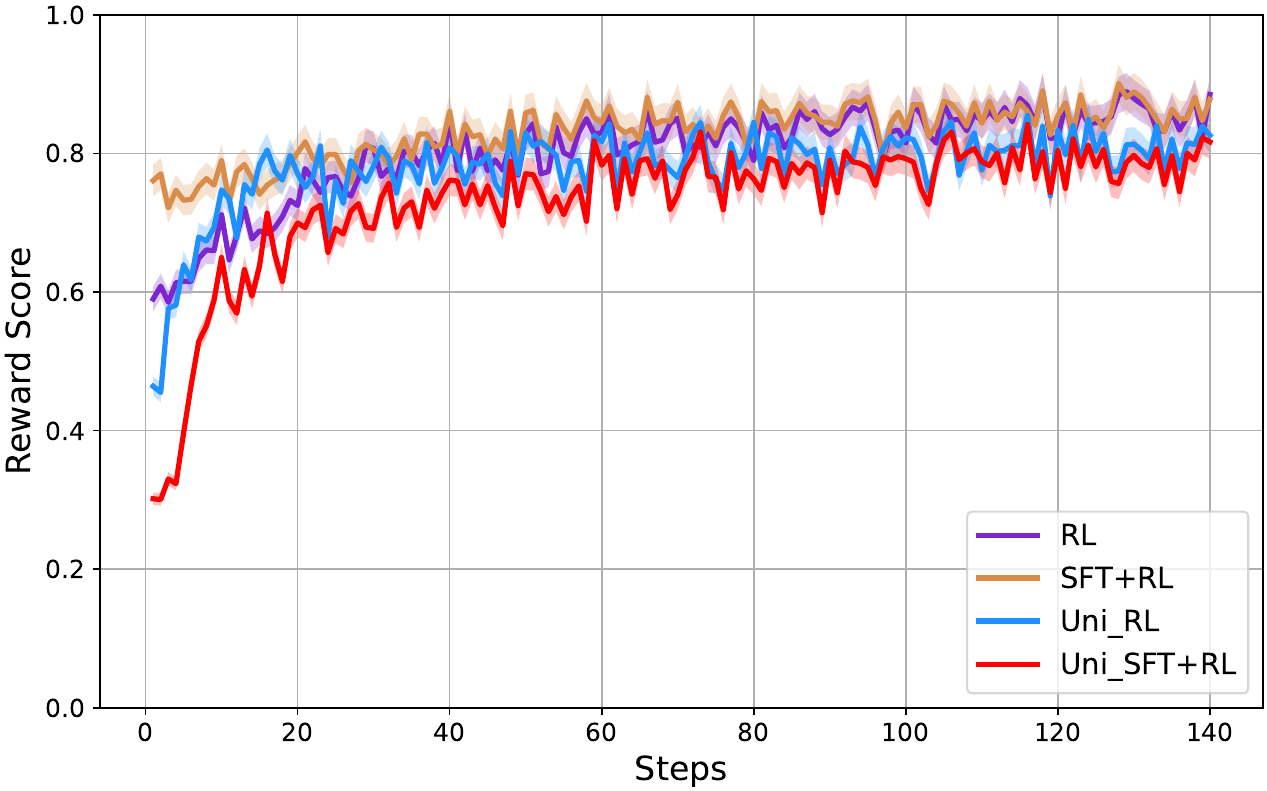}
        \caption{Reward with training steps.}
        \label{fig:reward_training}
    \end{subfigure}
    \hfill
    \begin{subfigure}[t]{0.32\textwidth}
        \centering
        \includegraphics[width=\textwidth]{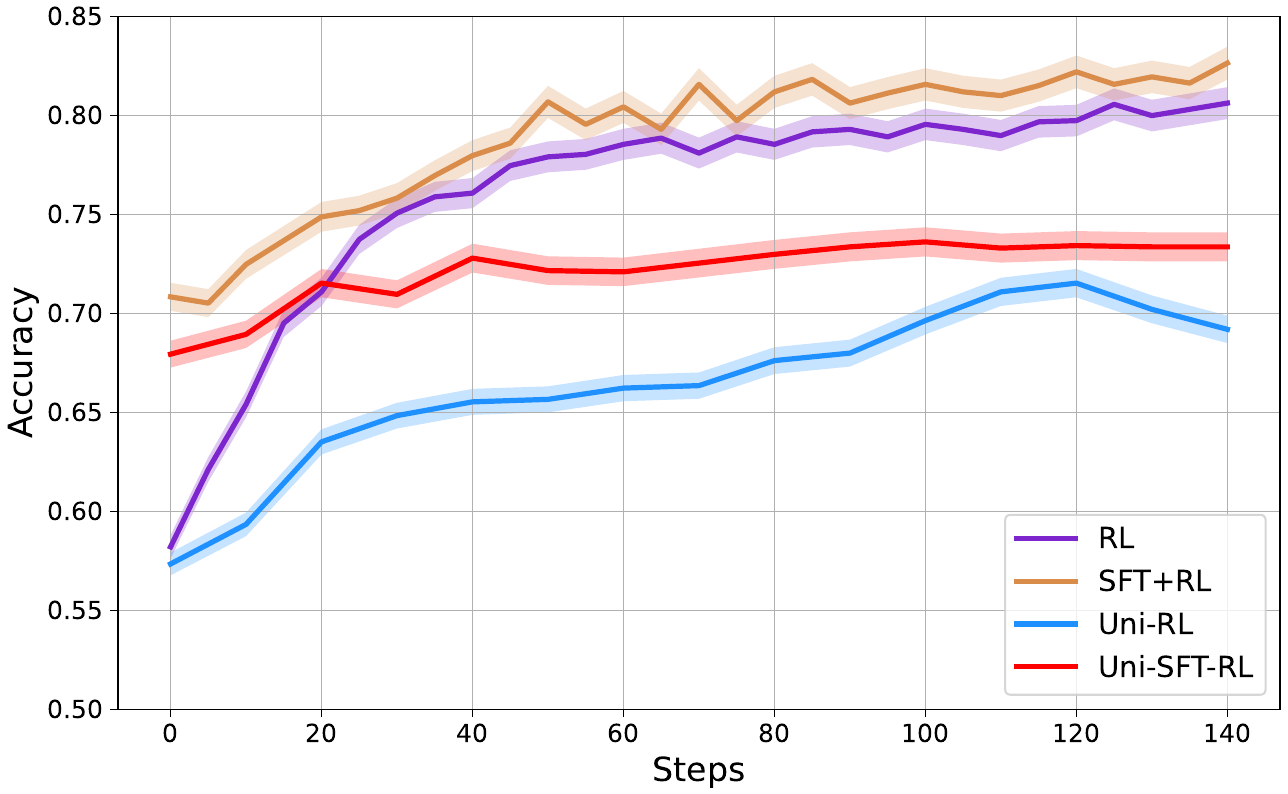}
        \caption{EM score with training steps.}
        \label{fig:score_training}
    \end{subfigure}
    \caption{We present the training metrics for the HiTab dataset, showing the performance of RL and SFT+RL under both the single-dataset setting and the unified-dataset setting.}
    \label{fig:hitab_training_metrics}
\end{figure*}

We also present the metric changes during the training process on the HiTab dataset. As shown in Figure \ref{fig:hitab_training_metrics}, the model trained with our carefully crafted reasoning data for cold start followed by RL training achieves the best performance.

\section{Prompts}

In this section, we present the prompts used in \ours. To ensure evaluation consistency, the prompt format is uniform across all LLMs. Table \ref{tab:reason_sft_generation} shows the reason-SFT reject sampling prompt. 
Table \ref{tab:train_prompt} shows the generation prompts. Table \ref{tab:position_instruction} presents the prompt for getting the position of tables. Table \ref{tab:llm_judge} shows the llm-as-a-judge prompt for LLM-based reward.

\begin{table}[h]
    \centering
    \caption{The prompt of reason-SFT generation. \textcolor{red}{question} and \textcolor{red}{table} will be replaced with the specific question and table.}
    \begin{tabular}{p{0.95\textwidth}} 
        \toprule
\#\# Objective: You are provided with a table and a question related to the table.\\
Your task is to answer the question step by step based on the table.\\
Table: \textcolor{red}{table} \\
Question: \textcolor{red}{question}

The answer should be short and simple. It can be a number, a word, or a phrase in the table, but not a full sentence.\\
Your response should end with `Answer: xxx` (answer to the question).\\
Now, answer the question step by step: \\
        \hline
    \end{tabular}
\label{tab:reason_sft_generation}
\end{table}

\begin{table}[h]
    \centering
    \caption{Input format of tableQA examples. \textcolor{red}{question} and \textcolor{red}{table} will be replaced. We use this template during training.}
    \begin{tabular}{p{0.95\textwidth}} 
        \toprule
Based on the table below, please answer the question, the answer should be short and simple. It can be a number, a word, or a phrase in the table, but not a full sentence.

\#\# Notes:

- Output the reasoning process inside a single pair of \think{...} tags.

- Output the final answer inside \answer{...} tags.

- Think through your reasoning process carefully, solve Question step by step.

\#\# Output Format Example

\think{Step-by-step reasoning here...}

\answer{here is your answer}

\#Task:

Table:
\textcolor{red}{table}

Question: 
\textcolor{red}{question} \\
        \hline
    \end{tabular}
\label{tab:train_prompt}
\end{table}

\begin{table}[h]
    \centering
    \caption{Prompt for getting the position of tables. \textcolor{red}{question} and \textcolor{red}{table} will be replaced with the specific question and table during training and inference. With two modes: \textit{(1) Column localization}: Identifying relevant columns, only use  \oneposition{column name}. \textit{(2) Column and Cell localization}: Identifying both columns and specific cells, use \position{cell content}{column name}. }
    \begin{tabular}{p{0.95\textwidth}} 
        \toprule
You are an expert in table-based reasoning. You are given a table and a question. \\
Your task is to answer the question based on the table.  

\# Guidelines \\
1. To answer the question, you should first think step-by-step by placing your reasoning trace inside \think{...} tags based on the table. \\
2. In your step-by-step reasoning inside \think{...} , ONLY annotate a table cell with \position{cell content}{column name} WHEN you are actually using that value in reasoning. Do NOT include or list all table cells—only those you are directly referencing. \\
3. Please use \answer{...} to enclose the final answer. The answer should be short and simple. It can be a number, a word, or a phrase in the table, but not a full sentence. It must be consistent with the question. \\
4. You must include \think{...} and \answer{...} tags in your response. \\

\# Task \\
Table: \textcolor{red}{table} \\
Question: \textcolor{red}{question} \\
        \hline
    \end{tabular}
\label{tab:position_instruction}
\end{table}

\begin{table}[h]
    \centering
    \caption{The prompt for the LLM-as-a-judge reward: Note that this reward is only used as a supplement and analysis for our method, and we don't use this reward on the full experiment.}
    \begin{tabular}{p{0.95\textwidth}} 
        \toprule
Here is the original question, the correct answer, and the candidate answer. Please evaluate whether the correct answer and the candidate answer are consistent. 

\# Examples:\\
- \\
Question: What is the distance from Paris to London? \\
Candidate Answer: 5 km\\
Correct Answer: 5000 m\\
Consistent: Yes\\
-\\
Question: How many people live in the city?\\
Candidate Answer: 1 million\\
Correct Answer: 1000000\\
Consistent: Yes\\
-\\
Question: What is the date today?\\
Candidate Answer: 2023-10-01\\
Correct Answer: October 1, 2023\\
Consistent: Yes\\
-\\
Question: What is the temperature in Paris?\\
Candidate Answer: 25°C\\
Correct Answer: 77°F\\
Consistent: No\\
-\\
Question: What is the distance from Paris to London?\\
Candidate Answer: 5 km\\
Correct Answer: 10 km\\
Consistent: No\\
-\\
\# YOUR TASK\\
Respond with only Yes or No. Do not include a rationale.\\
\texttt{Question: \{question\}}\\
\texttt{Candidate Answer: \{candidate\_answer\}}\\
\texttt{Correct Answer: \{correct\_answer\}} \\
\texttt{Consistent:} \\
        \hline
    \end{tabular}
\label{tab:llm_judge}
\end{table}

\section{Examples}
We demonstrate \ours’s reasoning process using representative examples from the training phase. Table~\ref{tab:evidence_case} traces a complete reasoning trajectory with explicit position annotations, revealing the provenance of each evidence fragment and the step-wise inference leading to the correct answer. Table~\ref{tab:wikitq_case} provides an example from the WikiTQ dataset, demonstrating the model’s ability to pinpoint and leverage critical table entries. Table~\ref{tab:tatqa_case} highlights a case from the TAT-QA benchmark, showcasing \ours’s proficiency in numerical calculation and internal consistency checks. In Table~\ref{tab:feverous_case}, we present a FEVEROUS example that demonstrates the model’s fact-verification capabilities over structured tabular data. Table~\ref{tab:hitab_case} presents a HiTab example, underscoring the model’s effectiveness in extracting relevant information from vertically oriented tables. Finally, Table~\ref{tab:hybridqa_case} shows a HybridQA case that combines tabular and textual inputs, illustrating \ours’s proficiency in integrating and reasoning over structurally diverse information sources.

\begin{table}[h]
\centering
\footnotesize
\caption{A case with position sources. \position{cell content}{column name} indicates position source.}
\label{tab:evidence_case}
\begin{tabular}{p{0.95\textwidth}} 
        \hline
        \textbf{Question}: braden gellenthien won the men's competition in 2006. who won the next year? \\
        \hline
        \textbf{Ground Truth}: Jedd Greschock \\
        \hline
        \textbf{Table}:   \\
        \textbar{} Year \textbar{} Location \textbar{} Men's Individual \textbar{} Women's Individual \textbar{} Men's Team \textbar{} Women's Team \textbar{} Mixed Team \textbar{} Ref \textbar{} \\ 
        \textbar{} 1996 \textbar{} Vaulx-en-Velin \textbar{} Franck Dauphin (FRA) \textbar{} Anna Campagnoli (ITA) \textbar{} None \textbar{} None \textbar{} None \textbar{} [226] \textbar{} \\
        \textbar{} 1998 \textbar{} Taoyuan \textbar{} John Blaschke (USA) \textbar{} Shih Ya-Ping (TPE) \textbar{} United States \textbar{} Chinese Taipei \textbar{} None \textbar{} [227] \textbar{} \\
        \textbar{} 2000 \textbar{} Madrid \textbar{} Tsai Sung-Pin (TPE) \textbar{} Jamie van Natta (USA) \textbar{} United States \textbar{} United States \textbar{} None \textbar{} [228] \textbar{} \\
        \textbar{} 2002 \textbar{} Chonburi \textbar{} Adam Wheatcroft (USA) \textbar{} Mary Zorn (USA) \textbar{} United States \textbar{} United States \textbar{} None \textbar{} [229] \textbar{} \\
        \textbar{} 2004 \textbar{} Madrid \textbar{} Choi Yong-Hee (KOR) \textbar{} Mary Zorn (USA) \textbar{} South Korea \textbar{} United States \textbar{} None \textbar{} [230] \textbar{} \\
        \textbar{} 2006 \textbar{} Viničné \textbar{} Braden Gellenthien (USA) \textbar{} Amandine Bouillot (FRA) \textbar{} United States \textbar{} France \textbar{} United States \textbar{} [231] \textbar{} \\
        \textbar{} 2008 \textbar{} Tainan \textbar{} Jedd Greschock (USA) \textbar{} Erika Anschutz (USA) \textbar{} Mexico \textbar{} South Korea \textbar{} United States \textbar{} [232] \textbar{}\\
        \textbar{} 2010 \textbar{} Shenzhen \textbar{} Adam Gallant (USA) \textbar{} Vida Halimian (IRI) \textbar{} United States \textbar{} United States \textbar{} United States \textbar{} [233] \textbar{}\\
        \textbar{} 2012 \textbar{} Córdoba \textbar{} Daniel Suter (USA) \textbar{} Kristina Berger (GER) \textbar{} Italy \textbar{} United States \textbar{} United States \textbar{} [234] \textbar{}\\
        \textbar{} 2014 \textbar{} Legnica \textbar{} None \textbar{} None \textbar{} None \textbar{} None \textbar{} None \textbar{} None \textbar{} \\
        \hline
        \textbf{Response}: \\
\think{Let me analyze this step by step:\\
1. The question states that Braden Gellenthien won the men's competition in 2006.\\
2. Looking at the table, I can verify that  \position{Braden Gellenthien (USA)}{Men's Individual} indeed won in \position{2006}{Year}.\\
3. The question asks who won the next year after 2006, which would be 2007.\\
4. Looking at the table, I don't see any entry for 2007. The years jump from 2006 to 2008.\\
5. In \position{2008}{Year}, I can see that \position{Jedd Greschock (USA)}{Men's Individual} won the men's individual competition.\\
So the next winner after Braden Gellenthien that appears in the table is Jedd Greschock in 2008. However, there is no data for 2007 in the table.\\
Wait, I need to be careful here. The question specifically asks about ``the next year'' which would be 2007. But since there's no 2007 entry in the table, I should clarify that the next year shown in the data after 2006 is 2008, when Jedd Greschock won.} \\
\hline
\answer{Jedd Greschock}\\
        \hline
    \end{tabular}
\end{table}

\begin{table}[h]
\centering
\footnotesize
\caption{A task example of WikiTQ.}
\label{tab:wikitq_case}
\begin{tabular}{p{0.95\textwidth}} 
        \hline
        \textbf{Question}: which stadium can hold more people than ballymena showgrounds, but less than windsor park? \\
        \hline
        \textbf{Ground Truth}: The Oval \\
        \hline
        \textbf{Table}: \\
\textbf{Table}: \\
\# \textbar{} Stadium \textbar{} Capacity \textbar{} City \textbar{} Home team \\
1 \textbar{} Windsor Park \textbar{} 24,734 \textbar{} Belfast \textbar{} Linfield, Northern Ireland national team \\
2 \textbar{} The Oval \textbar{} 15,000 \textbar{} Belfast \textbar{} Glentoran \\
3= \textbar{} Ballymena Showgrounds \textbar{} 8,000 \textbar{} Ballymena \textbar{} Ballymena United \\
3= \textbar{} New Grosvenor Stadium \textbar{} 8,000 \textbar{} Ballyskeagh \textbar{} Lisburn Distillery \\
3= \textbar{} Shamrock Park \textbar{} 8,000 \textbar{} Portadown \textbar{} Portadown \\
6 \textbar{} Brandywell Stadium \textbar{} 7,700 \textbar{} Derry \textbar{} Derry City \\
7= \textbar{} Seaview \textbar{} 6,500 \textbar{} Belfast \textbar{} Crusaders \\
7= \textbar{} The Showgrounds \textbar{} 6,500 \textbar{} Coleraine \textbar{} Coleraine \\
7= \textbar{} The Showgrounds \textbar{} 6,500 \textbar{} Newry \textbar{} Newry City \\
10= \textbar{} Inver Park \textbar{} 6,000 \textbar{} Larne \textbar{} Larne \\
10= \textbar{} Solitude \textbar{} 6,000 \textbar{} Belfast \textbar{} Cliftonville \\
13 \textbar{} Dixon Park \textbar{} 5,333 \textbar{} Ballyclare \textbar{} Ballyclare Comrades \\
14 \textbar{} Mourneview Park \textbar{} 5,000 \textbar{} Lurgan \textbar{} Glenavon \\
15 \textbar{} Donegal Celtic Park \textbar{} 4,200 \textbar{} Belfast \textbar{} Donegal Celtic \\
16= \textbar{} Holm Park \textbar{} 3,000 \textbar{} Armagh \textbar{} Armagh City \\
16= \textbar{} Lakeview Park \textbar{} 3,000 \textbar{} Loughgall \textbar{} Loughgall \\
16= \textbar{} Tillysburn Park \textbar{} 3,000 \textbar{} Belfast \textbar{} Harland \& Wolff Welders \\
21 \textbar{} Clandeboye Park \textbar{} 2,850 \textbar{} Bangor \textbar{} Bangor, Ards \\
22 \textbar{} Wilgar Park \textbar{} 2,500 \textbar{} Belfast \textbar{} Dundela \\
23= \textbar{} Allen Park \textbar{} 2,000 \textbar{} Antrim \textbar{} Chimney Corner \\
23= \textbar{} Hagan Park \textbar{} 2,000 \textbar{} Coagh \textbar{} Coagh United \\
23= \textbar{} Mill Meadow \textbar{} 2,000 \textbar{} Castledawson \textbar{} Wakehurst, Moyola Park \\
23= \textbar{} Riada Stadium \textbar{} 2,000 \textbar{} Ballymoney \textbar{} Glebe Rangers, Ballymoney United \\
28= \textbar{} Crystal Park \textbar{} 1,500 \textbar{} Banbridge \textbar{} Banbridge Town \\
28= \textbar{} Fortwilliam Park \textbar{} 1,500 \textbar{} Tobermore \textbar{} Tobermore United \\
28= \textbar{} Newforge Lane \textbar{} 1,500 \textbar{} Belfast \textbar{} PSNI \\
28= \textbar{} The Showgrounds (Limavady) \textbar{} 1,500 \textbar{} Limavady \textbar{} Limavady United \\
37 \textbar{} Glen Road Heights \textbar{} 400 \textbar{} Belfast \textbar{} Sport \& Leisure Swifts \\
38 \textbar{} Upper Braniel \textbar{} 300 \textbar{} Belfast \textbar{} Knockbreda \\
39 \textbar{} Milltown Playing Fields \textbar{} 280 \textbar{} Warrenpoint \textbar{} Warrenpoint Town \\
? \textbar{} Upper Malone \textbar{} ? \textbar{} Belfast \textbar{} Queen's University \\
        \hline
        \textbf{Response}: \\
\think{Okay, let's see. The question is asking for a stadium that can hold more people than Ballymena Showgrounds but less than Windsor Park. So first, I need to find the capacities of those two stadiums from the table.

Looking at the table, Ballymena Showgrounds (row 3=) has a capacity of 8,000. Windsor Park (row 1) has a capacity of 24,734. So we need stadiums with a capacity greater than 8,000 but less than 24,734.

Now, let's go through the list and check each stadium's capacity.

Stadium 2 is The Oval with 15,000. That's more than 8,000 and less than 24,734. So that's a candidate.

Stadium 3= has Ballymena Showgrounds at 8,000, so the next one after that is Stadium 6, Brandywell Stadium with 7,700. Wait, 7,700 is less than 8,000, so that's out.

Then Stadium 7= Seaview is 6,500, which is also under. The next is Stadium 10= Inver Park at 6,000. Still under. Solitude is 6,000 as well. Taylors Avenue is 6,000. Then Stadium 13 Dixon Park is 5,333. All those are below 8,000.

Stadium 14 Mourneview Park is 5,000. Then Stadium 15 Donegal Celtic Park is 4,200. The next set is from 16= onwards with 3,000, 2,850, etc. All those are way below 8,000.

So after Ballymena Showgrounds (8,000), the next stadium with a higher capacity is The Oval at 15,000. Then Stadium 6 is 7,700, which is below 8,000. So The Oval is the only one that's between 8,000 and 24,734. Wait, wait, let me check again.

Wait, The Oval is 15,000. That's definitely more than 8,000 and less than 24,734. So that's the answer. Are there any others? Let me check the list again.

Stadium 3= are all 8,000. Then Stadium 6 is 7,700 (below). Then the next stadiums after that are all below 8,000. So The Oval is the only one that fits. So the answer should be The Oval.} \\
\hline
\answer{The Oval}\\
        \hline
    \end{tabular}
\end{table}

\begin{table}[h]
\centering
\footnotesize
\caption{A task example of TAT-QA.}
\label{tab:tatqa_case}
\begin{tabular}{p{0.95\textwidth}} 
        \hline
        \textbf{Question}: What was the percentage change in equity settled share plans in 2019 from 2018? \\
        \hline
        \textbf{Ground Truth}: 8.77\% \\
        \hline
        \textbf{Table}:  \\ 
\textbar{} 2019 \textbar{} 2018 \\
\textbar{} £m \textbar{} £m \\
Adjusted operating profit \textbar{} 282.7 \textbar{} 264.9 \\
Depreciation and amortisation (excluding IFRS16 depreciation) \textbar{} 34.3 \textbar{} 32.9 \\
Cash payments to pension schemes in excess of charge to P\&L \textbar{} (5.2) \textbar{} (4.6) \\
Equity settled share plans \textbar{} 6.2 \textbar{} 5.7 \\
Working capital changes \textbar{} (21.4) \textbar{} (22.5) \\
Cash generation \textbar{} 296.6 \textbar{} 276.4 \\
        \hline
        \textbf{Response}: \\
\think{Okay, let's see. The question is asking for the percentage change in equity settled share plans from 2018 to 2019. Alright, so first, I need to find the values for those two years from the table.

Looking at the table, the row for Equity settled share plans has values of 6.2 for 2019 and 5.7 for 2018. So the 2018 value is 5.7 million pounds, and the 2019 value is 6.2 million.

To calculate the percentage change, the formula is [(New Value - Old Value)/Old Value] * 100. Here, the new value is 2019's 6.2, and the old value is 2018's 5.7.

Subtracting 5.7 from 6.2 gives 0.5. Then, divide that by the old value, 5.7. So 0.5 divided by 5.7. Let me compute that. 0.5 / 5.7 equals approximately 0.0877. Multiply by 100 to get the percentage, which is about 8.77\%.

Wait, let me check the calculation again. 0.5 divided by 5.7. Let's do 0.5 ÷ 5.7. 5.7 times 0.08 is 0.456, and 0.5 minus that is 0.044. So 0.044 divided by 5.7 is approximately 0.0077. So total is 0.0877, which is 8.77\%. Rounding to two decimal places, that's 8.77\%.

But sometimes percentage changes are rounded to one decimal. Let me confirm. The problem doesn't specify, so maybe keep two decimals. So the answer would be approximately 8.77\%. Let me make sure I didn't mix up the years. 2019 is the new value, 2018 the old. Yes, correctly. So the percentage increase is 8.77\%.} \\
\hline
\answer{8.77\%}\\
        \hline
    \end{tabular}
\end{table}

\begin{table}[h]
\centering
\footnotesize
\caption{A task example of FEVEROUS.}
\label{tab:feverous_case}
\begin{tabular}{p{0.95\textwidth}} 
        \hline
        \textbf{Claim}: Pescara Calcio a 5, which is based in Pescara, Italy, a city that was founded in 1811, won two games and lost six games in the 2015/2016 UEFA Futsal Cup. \\
        \hline
        \textbf{Ground Truth}: REFUTES \\
        \hline
        \textbf{Table}:   \\
Season \textbar{} Competition \textbar{} Round \textbar{} Opponent \textbar{} Result \\
2015/16 \textbar{} UEFA Futsal Cup \textbar{} Main Round \textbar{} Romania City'US Târgu Mureș \textbar{} Win 8–3 \\
- \textbar{} - \textbar{} - \textbar{} Germany Hamburg Panthers \textbar{} Win 11–0 \\
- \textbar{} - \textbar{} - \textbar{} Ukraine FC Lokomotiv Kharkiv \textbar{} Win 5–1 \\
- \textbar{} - \textbar{} Elite Round \textbar{} Belgium FP Halle-Gooik \textbar{} Win 4–2 \\
- \textbar{} - \textbar{} - \textbar{} Kazakhstan Tulpar Karagandy \textbar{} Win 4–0 \\
- \textbar{} - \textbar{} - \textbar{} North Macedonia KMF Zelezarec Skopje \textbar{} Win 2–0 \\
- \textbar{} - \textbar{} Final Four \textbar{} Spain Inter FS \textbar{} Loss 2–4 \\
- \textbar{} - \textbar{} - \textbar{} Portugal SL Benfica \textbar{} Loss (p) 2–2 (0–2) \\
2017/18 \textbar{} UEFA Futsal Cup \textbar{} Main Round \textbar{} Kazakhstan Kairat Almaty \textbar{} Loss 1–5 \\
- \textbar{} - \textbar{} - \textbar{} Belarus Stalitsa Minsk \textbar{} Loss 4–6 \\
- \textbar{} - \textbar{} - \textbar{} Russia Dinamo Moskva \textbar{} Win 5–3 \\
- \textbar{} - \textbar{} Elite Round \textbar{} Serbia Ekonomac \textbar{} Loss 2–3 \\
- \textbar{} - \textbar{} - \textbar{} Netherlands ZVV 't Knooppunt \textbar{} Win 9–1 \\
- \textbar{} - \textbar{} - \textbar{} Spain Barcelona \textbar{} Loss 1–3 \\
        \hline
        \textbf{Response}: \\
\think{Okay, let's try to figure out the correct label for this claim. The claim is about Pescara Calcio a 5 winning two games and losing six in the 2015\/2016 UEFA Futsal Cup. The table provided is for table\_1, which lists their competitions.

First, let me look at the 2015\/16 UEFA Futsal Cup season in the table. The main round shows they played Romania City'US Târgu Mureș and won 8-3. Then they played Germany Hamburg Panthers and won 11-0, and Ukraine FC Lokomotiv Kharkiv and won 5-1. That's three wins so far. Then in the Elite Round, they played Belgium FP Halle-Gooik and won 4-2, Kazakhstan Tulpar Karagandy 4-0, and North Macedonia KMF Zelezarec Skopje 2-0. That's three more wins. Then in the Final Four, they lost to Spain Inter FS 2-4 and Portugal SL Benfica, with a loss (p) 2-2 (0-2). So in the Final Four, they had two losses.

So tallying the wins: main round 3, elite round 3, total 6 wins. Losses: Final Four 2. But the claim says they won two games and lost six. Wait, that's the opposite of what the table shows. The table shows 6 wins and 2 losses in the 2015\/16 UEFA Futsal Cup. The claim states the opposite.

But wait, the user also mentioned the text cell (Pescara Calcio a 5 R1C1): Win 8–3. Looking at the table, the main round's first game against Romania City'US Târgu Mureș was a win 8-3. So maybe the text cell is referring to that same game? But the claim's numbers are reversed. If the actual results were 6 wins and 2 losses, but the claim says 2 wins and 6 losses, then that would be REFUTES.

Another thing to check: Does the text's mention of "Win 8–3" match any of their games? Yes, in the main round against Romania City'US Târgu Mureș, they did win 8-3. But the claim is about the total wins and losses in the season, not a single game. So even if the team did have some games with high scores, the overall record is six wins and two losses. Therefore, the claim is REFUTED by the table.} \\
\hdashline
\answer{REFUTES}\\
        \hline
    \end{tabular}
\end{table}

\begin{table}[h]
\centering
\footnotesize
\caption{A task example of HiTab.}
\label{tab:hitab_case}
\begin{tabular}{p{0.95\textwidth}} 
        \hline
        \textbf{Question}: what was the percentage of acquaintances to their victims among all females accused of sexual assault in 2015? \\
        \hline
        \textbf{Ground Truth}: 53.5 \\
        \hline
        \textbf{Table}: \\  Table title: violent crime perpetrated by females, by relationship of the accused to the victim, 2015  \\
Type of Violent Crime \textbar{} Current/Former Spouse \textbar{} \% \textbar{} Intimate Relationship \textbar{} \% \textbar{} Other Family \textbar{} \% \textbar{} Acquaintance \textbar{} \% \textbar{} Stranger \textbar{} \% \textbar{} Total \textbar{} \% \\
Homicide \textbar{} 10 \textbar{} 23.8 \textbar{} 0 \textbar{} 0.0 \textbar{} 16 \textbar{} 38.1 \textbar{} 13 \textbar{} 31.0 \textbar{} 3 \textbar{} 7.1 \textbar{} 42 \textbar{} 100.0 \\
Attempted Murder \textbar{} 9 \textbar{} 23.1 \textbar{} 10 \textbar{} 25.6 \textbar{} 6 \textbar{} 15.4 \textbar{} 8 \textbar{} 20.5 \textbar{} 6 \textbar{} 15.4 \textbar{} 39 \textbar{} 100.0 \\
Other Violations Causing Death \textbar{} 0 \textbar{} 0.0 \textbar{} 0 \textbar{} 0.0 \textbar{} 0 \textbar{} 0.0 \textbar{} 2 \textbar{} 40.0 \textbar{} 3 \textbar{} 60.0 \textbar{} 5 \textbar{} 100.0 \\
Sexual Assaults - All Levels \textbar{} 8 \textbar{} 4.0 \textbar{} 23 \textbar{} 11.6 \textbar{} 43 \textbar{} 21.7 \textbar{} 106 \textbar{} 53.5 \textbar{} 18 \textbar{} 9.1 \textbar{} 198 \textbar{} 100.0 \\
Other Sexual Violations \textbar{} 0 \textbar{} 0.0 \textbar{} 9 \textbar{} 10.1 \textbar{} 28 \textbar{} 31.5 \textbar{} 42 \textbar{} 47.2 \textbar{} 10 \textbar{} 11.2 \textbar{} 89 \textbar{} 100.0 \\
Sexual Violations Specific to Children \textbar{} 0 \textbar{} 0.0 \textbar{} 8 \textbar{} 12.9 \textbar{} 18 \textbar{} 29.0 \textbar{} 27 \textbar{} 43.5 \textbar{} 9 \textbar{} 14.5 \textbar{} 62 \textbar{} 100.0 \\
Sexual Violations Not Specific to Children \textbar{} 0 \textbar{} 0.0 \textbar{} 1 \textbar{} 3.7 \textbar{} 10 \textbar{} 37.0 \textbar{} 15 \textbar{} 55.6 \textbar{} 1 \textbar{} 3.7 \textbar{} 27 \textbar{} 100.0 \\
Assault - Level 3 - Aggravated \textbar{} 45 \textbar{} 16.7 \textbar{} 93 \textbar{} 34.4 \textbar{} 42 \textbar{} 15.6 \textbar{} 68 \textbar{} 25.2 \textbar{} 22 \textbar{} 8.1 \textbar{} 270 \textbar{} 100.0 \\
Assault - Level 2 - Weapon/Bodily Harm \textbar{} 1310 \textbar{} 22.9 \textbar{} 1476 \textbar{} 25.8 \textbar{} 985 \textbar{} 17.2 \textbar{} 1452 \textbar{} 25.3 \textbar{} 506 \textbar{} 8.8 \textbar{} 5729 \textbar{} 100.0 \\
Assault - Level 1 - Common \textbar{} 4227 \textbar{} 18.3 \textbar{} 4910 \textbar{} 21.3 \textbar{} 4400 \textbar{} 19.1 \textbar{} 7540 \textbar{} 32.7 \textbar{} 1993 \textbar{} 8.6 \textbar{} 23070 \textbar{} 100.0 \\
Other Assaults \textbar{} 11 \textbar{} 0.7 \textbar{} 13 \textbar{} 0.9 \textbar{} 18 \textbar{} 1.2 \textbar{} 439 \textbar{} 28.9 \textbar{} 1038 \textbar{} 68.3 \textbar{} 1519 \textbar{} 100.0 \\
Forcible Confinement\/Kidnapping \textbar{} 19 \textbar{} 13.7 \textbar{} 25 \textbar{} 18.0 \textbar{} 64 \textbar{} 46.0 \textbar{} 18 \textbar{} 12.9 \textbar{} 13 \textbar{} 9.4 \textbar{} 139 \textbar{} 100.0 \\
Robbery \textbar{} 1 \textbar{} 0.3 \textbar{} 18 \textbar{} 4.6 \textbar{} 9 \textbar{} 2.3 \textbar{} 109 \textbar{} 27.9 \textbar{} 254 \textbar{} 65.0 \textbar{} 391 \textbar{} 100.0 \\
Total \textbar{} 6227 \textbar{} 16.5 \textbar{} 7343 \textbar{} 19.5 \textbar{} 6401 \textbar{} 17.0 \textbar{} 13340* \textbar{} 35.3 \textbar{} 4426 \textbar{} 11.7 \textbar{} 37737 \textbar{} 100.0 \\
        \hline
        \textbf{Response}: \\
\think{Okay, let's try to figure this out. The question is asking for the percentage of acquaintances among all females accused of sexual assault in 2015. Looking at the table, I need to find the row that corresponds to sexual assault levels.

First, I'll scan through the "type of violent crime" column to find entries related to sexual assault. There are several entries here. Let me check each one. The first sexual-related crime is "sexual assaults - all levels". Then there's "other sexual violations", "sexual violations specific to children", and "sexual violations not specific to children". Then there's "assault..." which might not be sexual, so probably not relevant here. The question specifically mentions "sexual assault", so I should focus on "sexual assaults - all levels".

In that row, the columns are divided by relationships: current\/former spouse, intimate relationship, other family, acquaintance, stranger, and total. Each has a number and a percent. The question is about the percentage of acquaintances. So for "sexual assaults - all levels", the acquaintance percentage would be the value in that column.

Looking at the acquaintance column for that row: the number is 106, and the percent is 53.5. The total for that row is 198, which adds up (106 is half of 198). So the answer should be 53.5\% here. But wait, let me make sure there's no other sexual assault categories. The next entries are other sexual violations, which might be separate. But since the question says "sexual assault", maybe only the "sexual assaults - all levels" is the correct row. The other entries under sexual violations might be subcategories, so including them would change the total. But the question is about all females accused of sexual assault, so the main category here is "sexual assaults - all levels", and the percentage for acquaintances is 53.5\%.

So the answer should be 53.5\%.} \\
\hline
\answer{53.5}\\
        \hline
    \end{tabular}
\end{table}

\begin{table}[h]
\centering
\footnotesize
\caption{A case of HybridQA.}
\label{tab:hybridqa_case}
\begin{tabular}{p{0.95\textwidth}} 
        \hline
        \textbf{Question}: What is the team most recently managed by the 2012 manager of Real Garcilaso ? \\
        \hline
        \textbf{Ground Truth}: Alfonso Ugarte \\
        \hline
        \textbf{Table}:   \\
        Table Introduction: The 2012 Torneo Descentralizado de Fútbol Profesional (known as the 2012 Copa Movistar for sponsorship reasons) is the ninety-sixth season of Peruvian football. A total of 16 teams are competing in the tournament, with Juan Aurich as the defending champion. The Torneo Descentralizado began on February 19 and is scheduled to end on December 9, 2012.\\
Team \textbar{} Manager \textbar{} Captain \textbar{} Kit manufacturer \textbar{} Shirt sponsor \\
Alianza Lima \textbar{} José Soto \textbar{} Juan Jayo \textbar{} Nike \textbar{} \\
Cienciano \textbar{} Raul Arias \textbar{} Julio Garcia \textbar{} Aries \textbar{} Movistar \\
Cobresol \textbar{} Octavio Vidales \textbar{} Gregorio Bernales \textbar{} Loma 's \textbar{} Universidad Alas Peruanas \\
Inti Gas \textbar{} Edgar Ospina \textbar{} Nick Montalva \textbar{} Walon \textbar{} Inti Gas \\
José Gálvez \textbar{} Javier Arce \textbar{} Marco Ruiz \textbar{} Real \textbar{} SIDERPERU \/Megaplaza \\
Juan Aurich \textbar{} Franco Navarro \textbar{} Luis Guadalupe \textbar{} Walon \textbar{} Grupo Oviedo \\
León de Huánuco \textbar{} Jean Ferrari \textbar{} Carlos Zegarra \textbar{} Walon \textbar{} Roky 's \\
Melgar \textbar{} Julio Zamora \textbar{} Antonio Meza Cuadra \textbar{} Marathon \textbar{} \\
Real Garcilaso \textbar{} Freddy García \textbar{} Ramón Rodríguez \textbar{} Walon \textbar{} I-RUN \\
Sport Boys \textbar{} Jorge Espejo \textbar{} Jorge Huamán \textbar{} Triathlon \textbar{} Gobierno Regional del Callao \\
Sport Huancayo \textbar{} Wilmar Valencia \textbar{} Sergio Ibarra \textbar{} Manchete \textbar{} Caja Municipal de Huancayo \\
Sporting Cristal \textbar{} Roberto Mosquera \textbar{} Erick Delgado \textbar{} Umbro \textbar{} Cerveza Cristal \/ Volkswagen \\
Unión Comercio \textbar{} Mario Viera \textbar{} Héctor Sosa \textbar{} Real \textbar{} New Holland \\
Universidad César Vallejo \textbar{} Víctor Rivera \textbar{} Carlos Galván \textbar{} Walon \textbar{} Universidad César Vallejo \\
Universidad San Martín \textbar{} Angel Cappa \textbar{} Leao Butrón \textbar{} Umbro \textbar{} Herbalife \\
Universitario \textbar{} Nolberto Solano \textbar{} John Galliquio \textbar{} Umbro \textbar{} Radio Exitosa \\
        \hline
        \textbf{Related Text Information}:  
        Real\_Garcilaso: Cusco FC ( known as Real Garcilaso until 2019 ) is a Peruvian football club based in the city of Cusco .
Freddy\_García\_(football\_manager): Freddy Manuel Petroleo García Loayza ( born 22 November 1959 ) is a Peruvian football manager who most recently was the manager of Alfonso Ugarte in the Copa Perú .
Ramón\_Rodríguez\_(footballer): Ramón Rodríguez del Solar ( born 8 September 1977 in Pilcopata , Cuzco ) is a Peruvian footballer who plays as a striker for Santa Rosa in the Torneo Descentralizado . In his long career , Rodríguez has played for the likes of Cienciano , FBC Melgar , Total Clean , Deportivo Municipal , Sport Boys , Alianza Atlético , Inti Gas Deportes , and Cobresol and Real Garcilaso . His nicknamed is El Ratón ( The Mouse )
Walon\_Sport: Walon Sport , or simply Walon , is a multinational textile business founded in 1989 in Peru . It is one of the principal sportswear manufacturers in Peru , and it provides the kits for various association football clubs in the Liga 1 ( Peruvian first division ) . The company further expanded its production when it acquired deals with clubs in the Colombian Categoría Primera A. Walon Sport has also been the sponsor of the Peru national football team from 1998 to 2010 . \\
        \hline
        \textbf{Response}: \\
\think{Okay, let's see. The question is asking: What is the team most recently managed by the 2012 manager of Real Garcilaso?

First, I need to figure out who the manager of Real Garcilaso was in 2012. Looking at the table provided under the 2012 Torneo Descentralizado section, Real Garcilaso's manager is listed as Freddy García. So the 2012 manager of Real Garcilaso is Freddy García.

Next, I need to find out what team he most recently managed. The related text information includes a section about Freddy García. Let me check that. The text says: "Freddy\_García\_(football\_manager): ... most recently was the manager of Alfonso Ugarte in the Copa Perú." So according to this, after his time with Real Garcilaso in 2012, his most recent managerial position was with Alfonso Ugarte, and that was in the Copa Perú. The Copa Perú is a different league, probably a lower division or a regional league. Since the question is about the most recent team he managed, it would be Alfonso Ugarte. But wait, I need to make sure there's no other team mentioned after Real Garcilaso in his managerial history. The text states "most recently was the manager of Alfonso Ugarte", which implies that's the latest one before the information was written. The text might not list all his past teams, just the latest. So the answer should be Alfonso Ugarte.} \\
\hline
\answer{Alfonso Ugarte}\\
        \hline
    \end{tabular}
\end{table}

\end{document}